\documentclass{article}

\PassOptionsToPackage{numbers}{natbib}
\usepackage[preprint]{neurips_2026}

\usepackage[utf8]{inputenc} 
\usepackage[T1]{fontenc}    
\usepackage{hyperref}       
\usepackage{url}            
\usepackage{booktabs}       
\usepackage{amsfonts}       
\usepackage{nicefrac}       
\usepackage{microtype}      
\usepackage{xcolor}         
\usepackage{multirow}
\usepackage{graphicx}
\usepackage{subcaption}
\usepackage{amsmath}

\usepackage{algorithm}
\usepackage{algpseudocode}

\title{Synthetic but Not Realistic:\\The Evaluation Challenge in Generative Modelling for Structured Electronic Medical Records}

\author{Nicholas I-Hsien Kuo$^{1}$, \textbf{Blanca Gallego}$^{1}$,
\textbf{Louisa Jorm}$^{1}$,\\ 
$^{1}$Centre for Big Data Research in Health, the University of New South Wales, Sydney, Australia\\
\footnotesize{\textcolor{white}{*}}\\
Corresponding author: Nicholas I-Hsien Kuo (\texttt{n.kuo@unsw.edu.au})
}

\begin{document}

\maketitle

\begin{abstract}
Synthetic healthcare data are widely proposed as privacy-preserving substitutes for real patient data, yet their evaluation remains dominated by statistical similarity and predictive performance that do not reflect clinical validity. We introduce a multi-dimensional evaluation framework grounded in epidemiology, assessing descriptive fidelity, clinical utility, and structural validity, corresponding to descriptive, predictive, and causal questions. We evaluate four representative generative paradigms -- GAN-based, VAE-boosted, diffusion-based, and masked modelling -- using PRIME-CVD, a 50{,}000-person cohort with known ground-truth structure. While all models reproduce marginal distributions, none simultaneously preserve subgroup structure, effect estimates, and dependency structure. Notably, models with strong distributional fidelity can exhibit poor calibration and distorted relationships, leading to unreliable inference. These results show that current evaluation practices can overestimate synthetic data quality and motivate domain-informed assessment based on the ability to support valid clinical and scientific conclusions.
\end{abstract}

\section{Introduction}

Synthetic healthcare data for structured \emph{electronic medical records} (EMRs) are widely proposed as a solution to restricted access to individual-level clinical data, supporting data sharing~\cite{nicholas2024enriching}, model development and reproducibility~\cite{smith2023evaluating}, while addressing privacy and governance constraints~\cite{chen2021synthetic, befring2022transformation}. Advances in generative modelling have reinforced this promise, with models increasingly able to match real data in marginal distributions and predictive performance~\cite{figueira2022survey} in tabular settings~\cite{murtaza2023synthetic}.

However, the central challenge is not generation, but evaluation. Current practice is dominated by statistical similarity metrics and downstream predictive tasks, which provide limited insight into whether synthetic data support valid clinical or scientific use. A systematic review reports that over 80\% of studies rely primarily on these criteria~\cite{vallevik2024can}, with comparatively little attention to structural validity or subgroup behaviour. As a result, synthetic data may appear realistic under standard metrics yet fail to preserve properties required for reliable inference.

From an epidemiological perspective, this misalignment is fundamental. Healthcare research questions are typically descriptive~\cite{sun2014use}, predictive~\cite{hernan2006estimating}, or causal~\cite{tennant2021use}, each requiring preservation of distinct data characteristics. Descriptive analyses depend on accurate subgroup distributions, predictive tasks on well-calibrated risk estimates, and causal inference on correct dependency structure. Existing evaluation frameworks rarely distinguish between these requirements, implicitly assuming that improved statistical fidelity in variable distributions implies improved utility.

In this work, we propose a multi-dimensional evaluation framework grounded in these principles, operationalising descriptive fidelity, clinical utility, and structural validity. Using PRIME-CVD~\cite{kuo2026prime}, a cohort with known ground-truth structure, we evaluate four generative paradigms and show that while all models reproduce marginal distributions, none simultaneously preserve subgroup structure, effect estimates, and dependencies. These findings suggest that current evaluation practices overestimate synthetic data quality and motivate assessment based on their ability to support valid inference.

\section{Related work}
\subsection{Generative models for structured electronic medical records data}

As highlighted in a recent survey by \citet{vallevik2024can}, generative modelling for structured EMRs has primarily centred on four paradigms: variational autoencoders (VAEs)~\cite{kingma2013auto}, generative adversarial networks (GANs)~\cite{goodfellow2014generative, arjovsky2017wasserstein, gulrajani2017improved}, denoising diffusion probabilistic models (DDPMs)~\cite{sohl2015deep, ho2020denoising}, and masking-based autoencoding approaches~\cite{vincent2008extracting, pathak2016context}. Rather than competing families, these paradigms impose distinct inductive biases on how data distributions are learned and generated.

VAEs formulate generation as approximate Bayesian inference, optimising a variational lower bound to learn latent representations. Their strengths include stable training and efficient generation, but likelihood-based objectives often lead to oversmoothing and reduced fidelity~\cite{alemi2017deep, larsen2016autoencoding}. GANs instead rely on an adversarial objective~\cite{schmidhuber2020generative}, enabling high-fidelity synthesis without explicit likelihood modelling. However, they are prone to instability, hyperparameter sensitivity, and reduced diversity due to mode collapse~\cite{arjovsky2017wasserstein, goodfellow2016nips}, challenges that are amplified in heterogeneous clinical data~\cite{kuo2022health}.

Diffusion and score-based models~\cite{sohl2015deep, ho2020denoising} adopt a stochastic process perspective, learning to iteratively denoise corrupted data. These approaches offer strong mode coverage and training stability~\cite{dhariwal2021diffusion}, but require many sampling steps, limiting efficiency in real-time healthcare applications~\cite{song2020score}. Masking-based autoencoding models reconstruct partially observed inputs~\cite{devlin2019bert}, naturally aligning with EMR data where missingness and feature dependencies are intrinsic (\textit{e.g.,} BEHRT~\cite{li2020behrt}). While effective for representation learning and imputation, they are less suited for unconditional generation.

Despite these advances, generative modelling in healthcare remains relatively new when compared to vision and language domains. More recent paradigms such as rectified flow models~\cite{liu2022flow} and hybrid approaches (\textit{e.g.,} VAE-GAN~\cite{larsen2016autoencoding}, latent diffusion~\cite{rombach2022high}) remain underexplored in EMR settings, despite offering potential trade-offs between fidelity, stability, and efficiency. In parallel, the focus in clinical applications has increasingly shifted from generative performance to robustness, fairness, and evaluation, reflecting the high-stakes nature of healthcare deployment~\cite{mcgreivy2024weak}.

\subsection{Evaluation of synthetic healthcare data}

Evaluation of synthetic EMR data has largely focused on statistical similarity and downstream utility~\cite{figueira2022survey}. Statistical similarity is assessed through univariate and multivariate comparisons, using metrics such as Kolmogorov--Smirnov tests~\cite{an1933sulla, smirnov1948table}, divergence measures (Kullback--Leibler~\cite{kullback1951information}, Jensen--Shannon~\cite{dagan1997similarity}), and integral probability metrics including Maximum Mean Discrepancy~\cite{song2008learning} and Wasserstein distance~\cite{givens1984class}. Visual techniques such as PCA~\cite{gewers2021principal} and t-SNE~\cite{van2008visualizing} are commonly used for qualitative inspection, but lack formal guarantees.

Beyond marginal distributions, recent work has emphasised structural fidelity. Methods based on propensity scores~\cite{rosenbaum1983central} and clustering~\cite{ybanez2023can} evaluate overlap between real and synthetic samples, capturing discrepancies in latent structure not detectable through marginal metrics~\cite{pathare2023comparison}. Utility-based evaluation assesses downstream performance by training models on synthetic data and testing on real data, often reporting comparable predictive accuracy~\cite{wang2021generating, dankar2022multi}. However, such utility does not guarantee fidelity.

Recent studies explicitly distinguish these concepts. \citet{hernadez2023synthetic} separate resemblance from utility, while \citet{alaa2022faithful} decompose data quality into fidelity, diversity, and generalisation. In healthcare, evaluation is frequently framed around privacy risks such as re-identification~\cite{el2020evaluating}, leading to an emphasis on security over realism. However, emerging evidence suggests that current generative models lack sufficient fidelity for such risks to be practically consequential and remain inadequate substitutes for real clinical data~\cite{kuo2024ck4gen}.

Crucially, existing evaluation approaches remain misaligned with how synthetic EMR data are used in practice. In this work, we adopt an epidemiological perspective and assess three fundamental aspects -- descriptive, predictive, and causal -- operationalised through our framework (Section~\ref{Sec:3Stream}).

\newpage
\section{Methods}
\subsection{Dataset and problem setup}\label{sec:PrimeCVD}

\begin{table}[h!]
\caption{Summary characteristics of the PRIME-CVD cohort (N = 50{,}000).}
\centering
\renewcommand{\arraystretch}{1.2}
\begin{tabular}{lll}
\toprule
\textbf{Variable} & \textbf{Statistic} & \textbf{Value} \\
\hline\hline
Age (years) & Mean (SD) & 49.71 (12.37) \\
IRSD Quintile (\%) & Q1--Q5 & 21.28 / 16.11 / 23.88 / 16.99 / 21.74 \\
Smoking Status (\%) & Non / Ex / Current & 73.14 / 16.72 / 10.13 \\
\hline
BMI (kg/m$^2$) & Mean (SD) & 28.33 (5.03) \\
SBP (mmHg)     & Mean (SD) & 123.31 (16.10) \\
eGFR (mL/min/1.73m$^2$) & Mean (SD) & 82.77 (6.09) \\
HbA1c (\%)     & Mean (SD) & 4.79 (0.93) \\
\hline
Disease prevalence (\%) & Diabetes / CKD / AF & 7.43 / 0.68 / 0.72 \\
\hline
5-year CVD event & Rate (\%) & 4.02 \\
Follow-up time   & Mean (years) & 4.80 \\
\bottomrule
\end{tabular}
\label{Table:Table01}
\end{table}

PRIME-CVD (Table~\ref{Table:Table01}) is a simulated cohort of 50{,}000 adults generated \emph{de novo} from a hand-specified directed acyclic graph (DAG) parameterised using publicly available Australian epidemiologic statistics~\cite{kuo2026prime, nicholas2025estimating}. It represents a primary prevention population with no prior CVD and simulates 5-year cardiovascular risk. The dataset integrates mixed-type, multi-modal variables spanning demographics, socioeconomic status (index of relative socio-economic disadvantage [IRSD]~\cite{walker2005index}; quintile 1 is most socioeconomically disadvantaged), chronic conditions, and biomarkers. All variables are generated through explicit structural relationships (Figure~\ref{fig:4DAGs}), enabling rigorous evaluation of statistical, clinical, and causal properties.  The dataset is publicly available under a CC-BY 4.0 licence~\cite{Kuo2026}.

\paragraph{Unified representation.}
Each sample is encoded as
\(
x = (x^{(r)}, x^{(1)}, \dots, x^{(K)}) \in \mathcal{X},
\)
where\\\(x^{(r)} \in [0,1]^{d_r}\) are continuous variables and \(x^{(k)} \in \Delta^{C_k-1}\) are one-hot categorical variables. Continuous variables are transformed via a Box--Cox transformation followed by min--max scaling to \([0,1]\), while categorical variables are mapped to one-hot representations. The resulting data space is
\[
\mathcal{X} = [0,1]^{d_r} \times \prod_{k=1}^K \Delta^{C_k-1} \subset \mathbb{R}^d,
\quad d = d_r + \sum_{k=1}^K C_k,
\]
with \(P_r\) denoting the empirical distribution.

\paragraph{Generative framework.}
All models $\mathcal{M}$ produce synthetic samples via
\(
\tilde{x} = \mathcal{M}_\theta(u), \quad \tilde{x} \in \mathcal{X}.
\)\\
In all cases, \(\mathcal{M}_\theta\) with parameter $\theta$ is trained to approximate the data distribution \(P_r\) under a common representation, ensuring generated samples are directly comparable across paradigmss.

\paragraph{Post-processing.}
Model outputs $\tilde{x}$ are mapped to the original domain by inverting preprocessing: continuous variables are rescaled and inverse Box--Cox transformed, while categorical variables are recovered via $\arg\max$ over simplex blocks and mapped to labels. Despite differing architectures, all models operate on the same space $\mathcal{X}$, enforce schema constraints (sigmoid/softmax), and aim to learn $P_\theta \approx P_r$, enabling consistent evaluation of marginal distributions and dependency structure.

\subsection{Generative models}\label{sec:models}

We consider four generative paradigms for tabular synthesis under the unified mapping, differing only in input $u$: (i) $z \sim \mathcal{U}$ (WGAN-GP), (ii) $z \sim p(z)$ learned via VAE (WGAN-GP+VAE), (iii) $x_T \sim \mathcal{N}(0,I)$ with iterative denoising (DDPM), and (iv) $x^{(0)} \sim P_{\text{ref}}$ with masked updates (MCM).

\paragraph{WGAN-GP.}
In \cite{kuo2022health}, a generator maps latent noise \(z \sim \mathcal{U}([0,1]^{d_z})\) to samples \(\tilde{x} = G_\theta(z)\). A critic $D_{\phi}$ approximates the Wasserstein distance under a gradient penalty (GP) enforcing 1-Lipschitzness:
\[
\min_\theta \max_{\phi \in \text{Lip}_1}
\mathbb{E}_{x \sim P_r}[D_\phi(x)]
-
\mathbb{E}_{z}[D_\phi(G_\theta(z))]
+
\lambda_{gp} \mathcal{L}_{GP}.
\]
To preserve inter-variable structure, the generator objective includes a correlation regulariser\\
\(
\lambda_c \|C_g - C_r\|_1,
\)
where \(C_r\) and \(C_g\) denote correlation matrices of real and generated samples.

\paragraph{WGAN-GP + VAE.}  
In \cite{nicholas2023generating}, WGAN-GP is extended by replacing the fixed prior with a learned latent distribution. A VAE encoder defines  
\[
q(z|x) = \mathcal{N}(\mu(x), \mathrm{diag}(\sigma^2(x))),
\]  
and sampling is performed from an empirical latent distribution induced by the encoder. In practice, this is implemented via a finite buffer of recent latent means, which are updated online and subsampled during generation. This buffer provides a Monte Carlo approximation to the aggregated posterior $\mathbb{E}_{x}[q(z|x)]$ without explicitly fitting a parametric prior. The latent scale is tracked through an exponentially smoothed standard deviation,
\[
\tilde{\sigma} \leftarrow \alpha \tilde{\sigma} + (1-\alpha)\,\mathrm{Std}_{\text{batch}}(\mu),
\]
yielding the sampling rule $z = \mu_k + \epsilon \odot \tilde{\sigma}$, where $\mu_k$ is drawn from the buffer and $\epsilon \sim \mathcal{N}(0, I)$. This construction allows the generator input to avoid a fixed Gaussian assumption. The generator $\tilde{x} = G_\theta(z)$ uses a Transformer encoder~\cite{vaswani2017attention}, while the critic and objective follow WGAN-GP.

\paragraph{Diffusion (DDPM).}
In \cite{nicholas2023synthetic}, samples are generated by reversing a Gaussian noising process. Starting from \(x_T \sim \mathcal{N}(0,I)\), the model iteratively denoises via
\[
x_t = \sqrt{\bar{\alpha}_t} x_0 + \sqrt{1-\bar{\alpha}_t}\,\epsilon,
\quad
\epsilon_\theta(x_t,t) \approx \epsilon,
\]
where \(\bar{\alpha}_t = \prod_{s=1}^t \alpha_s\) denotes the cumulative product of noise schedule coefficients where $\alpha_s\in(0, 1)$. An estimate of the clean sample is
\[
\hat{x}_0 = \frac{x_t - \sqrt{1-\bar{\alpha}_t}\,\epsilon_\theta(x_t,t)}{\sqrt{\bar{\alpha}_t}},
\]
and the final output is \(\tilde{x} = \Pi(\hat{x}_0)\), where \(\Pi\) enforces schema constraints. The training objective combines diffusion, reconstruction, and structural terms:
\[
\mathcal{L} =
\mathbb{E}\|\epsilon_\theta(x_t,t)-\epsilon\|_2^2
+
\lambda_r \|\Pi(\hat{x}_0)-x_0\|_2^2
+
\lambda_c \|\mathrm{Corr}(\Pi(\hat{x}_0)) - C_r\|_1.
\]

\paragraph{Masked Conditional Model (MCM).}
Instead of latent sampling, in \cite{nicholas2025attention}, MCM learns conditional reconstruction. Given a reference sample \(x^{(0)} \sim P_{\text{ref}}\) and mask \(m\), the model predicts
\[
\hat{x} = f_\theta(x_m, m), \quad x_m = x \odot m + c \odot (1-m),
\]
where \(c\) is a learned attention mechanism. Training minimises masked reconstruction loss with correlation regularisation:
\[
\mathcal{L} = \mathcal{L}_{\text{recon}} + \lambda_c \|\tilde{C}_f - \tilde{C}_r\|_1.
\]
Generation proceeds iteratively, producing a final sample \(\tilde{x} = x^{(R)}\):
\[
x^{(t+1)} = m^{(t)} \odot x^{(t)} + (1-m^{(t)}) \odot f_\theta(x^{(t)}, m^{(t)}).
\]

\paragraph{Supplementary details.}
Reproducibility details are provided in Appendix \S~\ref{App:Repro_Dependencies} (Python dependencies~\cite{van1995python}) and \S~\ref{App:Repro_CodeAvail} (code availability). Complete specifications of model architectures, training objectives, and hyperparameters for all four approaches are given in Appendices \S~\ref{App:Model1_All}--\ref{App:Model4_All}.

\subsection{Evaluation framework}\label{Sec:3Stream}

Epidemiological research fundamentally spans descriptive, predictive, and causal questions~\cite{carlin2026identifying, dyer2025distinction}; we evaluate synthetic data based on its readiness to substitute real data in supporting these use cases.

\paragraph{Descriptive fidelity.}
We assess whether synthetic data reproduce observable statistical properties of the real data. Following existing research, this is performed primarily through visual comparison of marginal distributions. 

To detect subgroup-specific distortions, we further examine stratified summaries across predefined subpopulations (\textit{e.g.,} IRSD quintiles and age groups), evaluating whether conditional distributions $P_\theta(X \mid H=h)$ align with $P_r(X \mid H=h)$. This ensures that apparent population-level realism is not driven by averaging across heterogeneous subgroups.

\paragraph{Clinical utility.}
PRIME-CVD is a time-to-event dataset~\cite{kalbfleisch2002statistical}, where each individual is represented by $(T, \delta, X)$, with event time $T$, event indicator $\delta \in \{0,1\}$, and covariates $X$. The primary quantity 

\newpage
of interest is the hazard function:
\begin{equation*}
\lambda(t \mid X) = \lim_{\Delta t \to 0} \frac{\mathbb{P}(t \leq T < t+\Delta t \mid T \geq t, X)}{\Delta t},    
\end{equation*}
which characterises instantaneous event risk over time.

We model this downstream clinical use case using the Cox proportional hazards model~\cite{cox1972regression}:
\begin{equation*}
\lambda(t \mid X) = \lambda_0(t)\exp(\beta^\top X),
\end{equation*}
where $\lambda_0(t)$ is an unspecified baseline hazard and $\beta$ are log-risk coefficients. This semi-parametric formulation enables estimation of covariate effects without imposing assumptions on baseline risk, making it standard for population health studies.

We fit Cox models on both real data $D_{\mathrm{real}}$ and sythetic data $D_{\mathrm{syn}}$, yielding coefficients $\hat{\beta}_{\mathrm{real}}$ and $\hat{\beta}_{\mathrm{syn}}$. The corresponding hazard ratios are $\mathrm{HR}_j = \exp(\beta_j)$, which represent multiplicative changes in risk associated with covariate $X_j$. Preservation of hazard ratios is critical, as they quantify effect sizes used for risk stratification, clinical interpretation, and causal inference.

Beyond relative effects, we assess calibration of predicted risk~\cite{van2019calibration}. For a fixed time horizon $\tau$ (defaulted as 5 years), the Cox model induces predicted risk $\hat{F}(\tau \mid X)$, representing absolute event probability. Calibration evaluates whether predicted risks match observed event frequencies, supporting accurate risk stratification and efficient resource use. Both real-trained and synthetic-trained models are evaluated on $D_{\mathrm{real}}$, and calibration curves are constructed by comparing predicted and observed event rates across bins. Calibration is summarised via the distance-to-1 metric (D21, the lower the better)~\cite{kuo2024ck4gen}:
\begin{equation*}
\mathrm{D21} = \left| \mathrm{slope} - 1 \right|,
\end{equation*}
where a slope of 1 indicates perfect agreement. Subgroup calibration is additionally evaluated to assess whether synthetic data preserve risk estimation consistently across heterogeneous populations.

\paragraph{Structural validity.}
We assess preservation of higher-order dependency structure using causal discovery, reflecting the need to maintain relationships beyond pairwise associations. Greedy Equivalence Search (GES)~\cite{chickering2002learning} is applied to both $D_{\mathrm{real}}$ and $D_{\mathrm{syn}}$, yielding directed edge sets $E^\ast$ and $\hat{E}$ over the covariates, corresponding to graphs learned from real and synthetic data, respectively.

Structural agreement is evaluated using adjacency and orientation precision, recall, and F1 scores, alongside Structural Hamming Distance (SHD, the lower the better)~\cite{peters2013structural}:
\begin{equation*}
\mathrm{SHD}(E^\ast, \hat{E}) = \sum_{i<j} d_{ij},
\end{equation*}
where $d_{ij}$ is an indicator of discrepancy for node pair $(i,j)$, taking value 1 if the edge differs in presence or direction between $E^\ast$ and $\hat{E}$, and 0 otherwise. These metrics jointly assess recovery of both graph skeleton and edge direction.

\paragraph{Supplementary details.}
Additional evaluation details are provided in Appendices \S~\ref{App:InfStream1}--\ref{App:InfStream3}, including definitions of the Cox model and metrics for clinical utility and structural validity.

\section{Results}\label{Sec:Results}
\begin{figure}[h!]
    \centering
    \includegraphics[width=\linewidth]{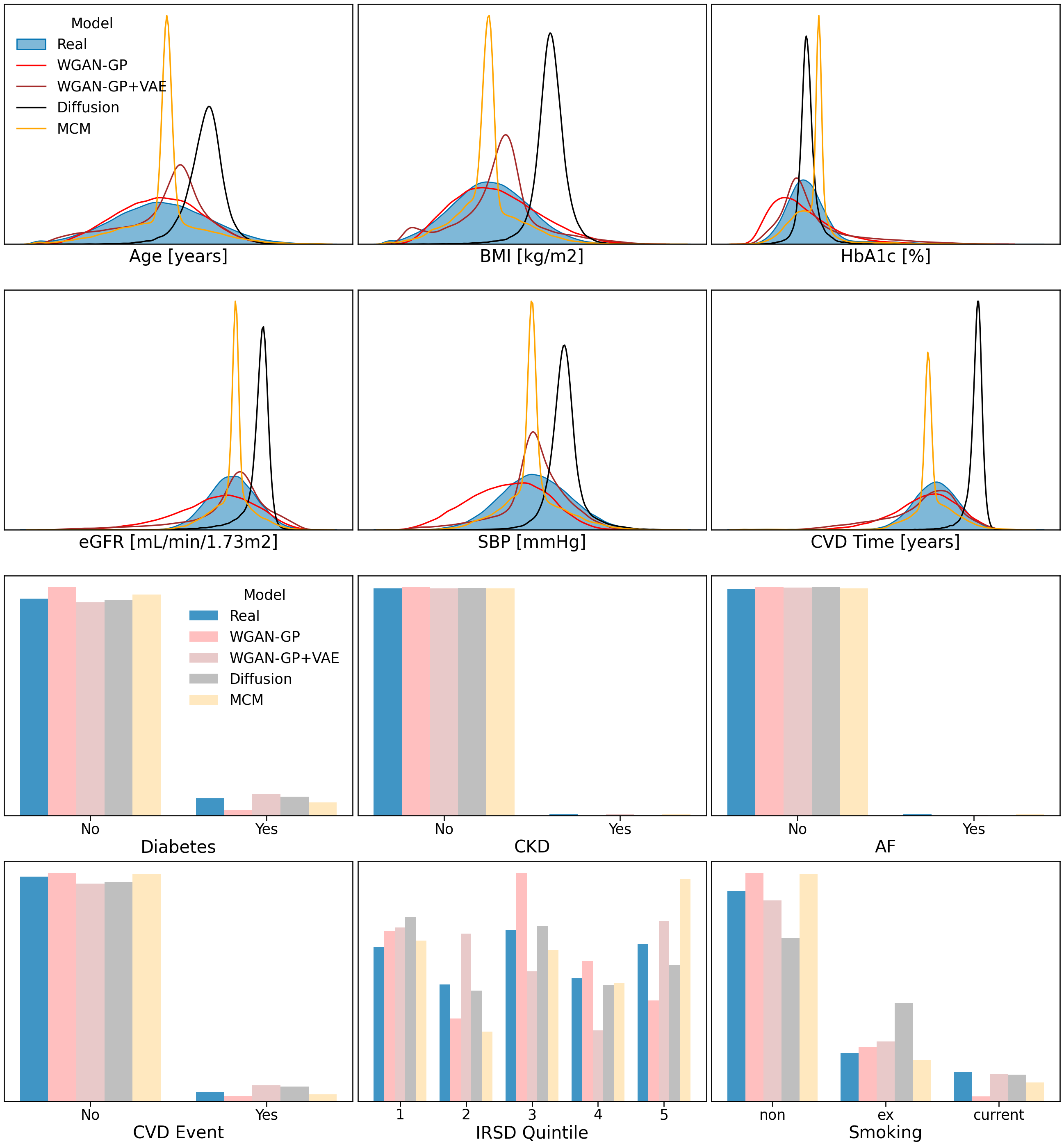}
    \caption{Distributional Fidelity of Synthetic PRIME-CVD Cohorts Across Generative Models}
    \label{fig:4Dist}
\end{figure}

\paragraph{Marginal distributions at population level.}
Figure~\ref{fig:4Dist} shows all models are capable to reproduce overall distributions consistent with the ground truth. Notably for continuous variables, DDPM and MCM exhibit reduced variance, with sharper and more concentrated distributions (\textit{e.g.,} BMI, SBP).

\begin{figure}
    \centering
    
    \begin{subfigure}[t]{0.99\linewidth}
        \centering
        \includegraphics[width=\linewidth]{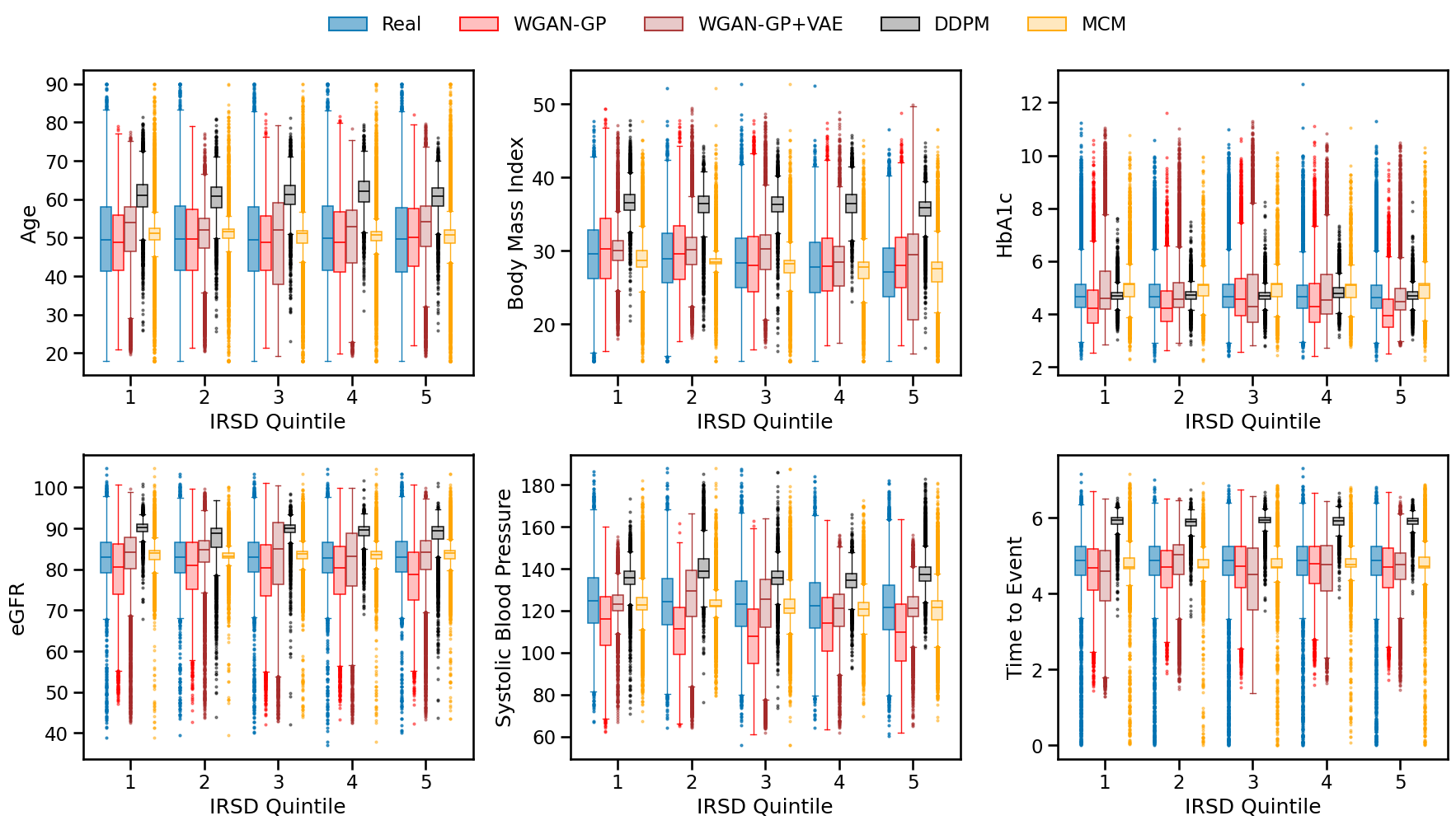}
        \caption{Numeric variable comparison across IRSD quintiles}
        \label{fig:4DistStratIRSD1}
    \end{subfigure}
    
    \vspace{0.5em}
    
    \begin{subfigure}[t]{0.985\linewidth}
        \centering
        \includegraphics[width=\linewidth]{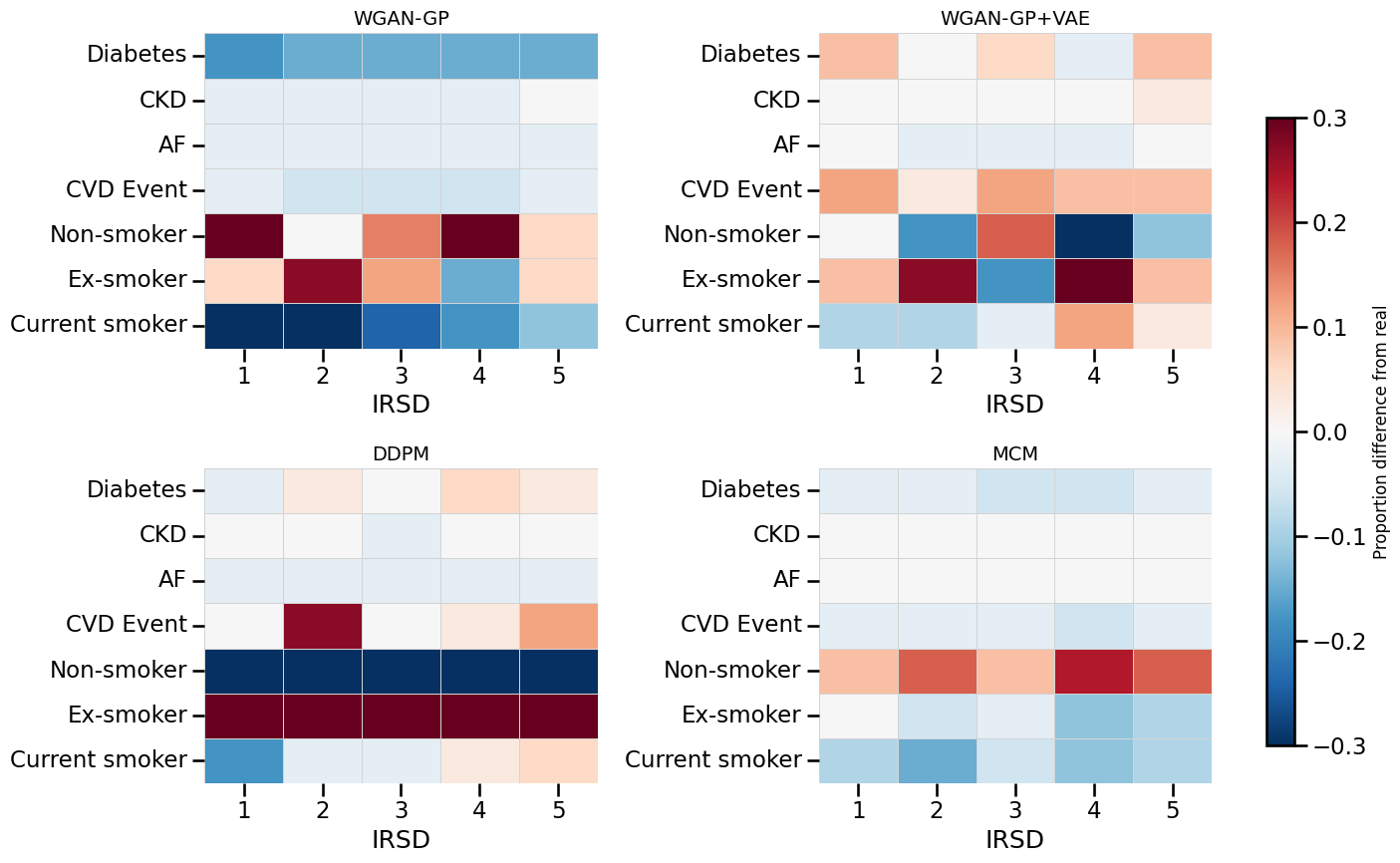}
        \caption{Categorical proportional differences across IRSD quintiles}
        \label{fig:4DistStratIRSD2}
    \end{subfigure}
    
    \caption{Distributional comparisons across IRSD quintiles for numeric and categorical variables.}
    \label{fig:4DistStratIRSD}
\end{figure}

\paragraph{Marginal distributions at subgroup level.}
Figure~\ref{fig:4DistStratIRSD} evaluates distributions conditional on IRSD quintiles. Continuous variable results (panel~(a)) mirror Figure~\ref{fig:4Dist}, with DDPM and MCM show reduced interquartile ranges. For categorical variables (panel~(b)), we report the proportional deviations from real data (\textit{e.g.,} $\mathbb{P}(\text{Diabetes}_\text{syn}) - \mathbb{P}(\text{Diabetes}_\text{real})$). All models exhibit increased deviations relative to the population level; with consistent directional shifts across IRSD quintiles.

\begin{figure}
    \centering
    
    \begin{subfigure}[t]{\linewidth}
        \centering
        \includegraphics[width=\linewidth]{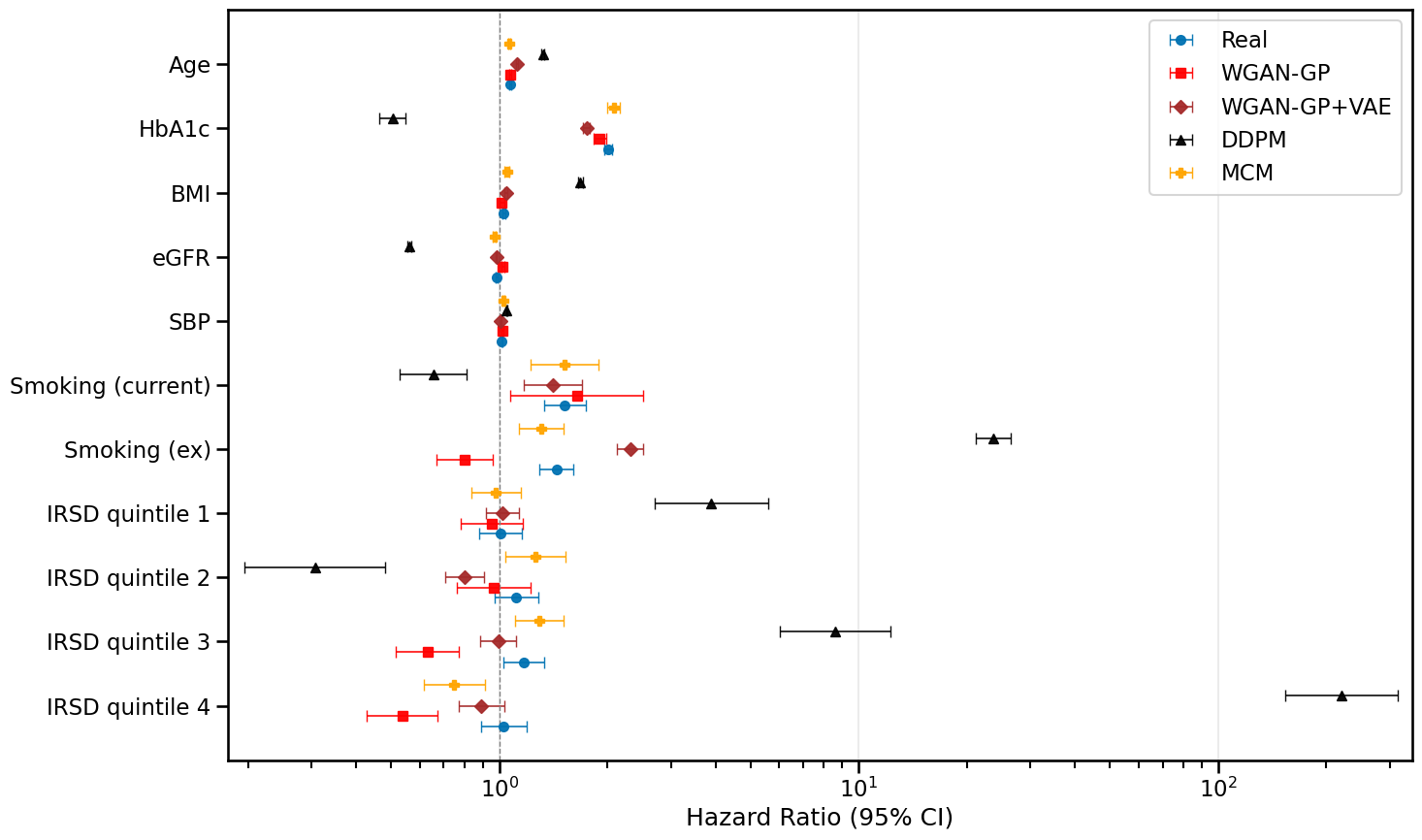}
        \caption{Clinical utility comparison using hazard ratios (log-scaled)}
        \label{fig:4DistHR}
    \end{subfigure}
    
    \vspace{0.5em}
    
    \begin{subfigure}[t]{\linewidth}
        \centering
        \includegraphics[width=\linewidth]{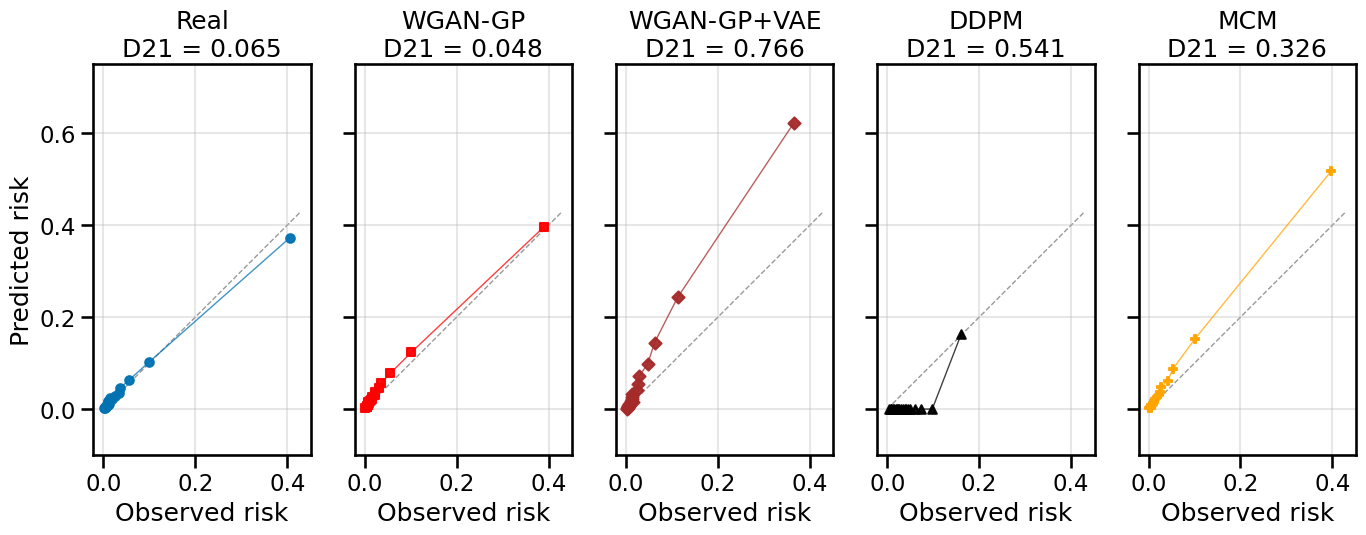}
        \caption{Clinical utility comparison using calibration curves (trained on synthetic data, tested on real data)}
        \label{fig:4Cali}
    \end{subfigure}
    
    \caption{Evaluation of clinical utility for models trained on real and synthetic data.}
    \label{fig:4ClinicalUtility}
\end{figure}

\paragraph{Hazard ratios.}
Figure~\ref{fig:4ClinicalUtility}(a) compares HRs estimated from Cox models. GAN-based methods and MCM produce HRs closest to those derived from real data, with overlapping confidence intervals in several covariates. In contrast, DDPM produces substantially larger HRs for multiple covariates.

\newpage

\paragraph{Calibration.}
Figure~\ref{fig:4ClinicalUtility}(b) shows calibration curves evaluated on real data at the population level, with D21 summarised in Table~\ref{tab:4Cali}. At the cohort level, synthetic data derived from WGAN-GP achieves the lowest deviation (D21 = 0.048), compared to 0.065 for the real model. Stratified calibration results (Table~\ref{tab:4Cali}) show increased deviation across IRSD quintiles for all models. For example, synthetic data derived from WGAN-GP ranges from 0.15 to 0.35 across IRSD quintiles. Synthetic data derived from other generative models exhibit larger deviations.

\paragraph{Graph structure.}
Figure~\ref{fig:4DAGs} shows directed graphs recovered using GES. MCM's synthetic data produces sparser graphs with fewer edges; while aternative synthetic data yield densely connected graphs. Across all models, no recovered graph identifies both Age and IRSD as exogenous nodes.

\paragraph{Quantitative structural metrics.}
Structural metrics across 20 runs are summarised in Table~\ref{Tab:4DAGsStability} using median and interquartile range. MCM achieves the lowest SHD (16 [11--21]), indicating the closest recovery of the reference graph, while DDPM shows the largest structural deviation (32 [25--35]).

For adjacency, MCM attains the highest F1 score (0.62 [0.48--0.69]), with relatively balanced precision and recall. In contrast, WGAN-GP, WGAN-GP+VAE, and DDPM exhibit higher recall (0.58--0.83)

\newpage

\newpage

\begin{table}[h!]
\centering
\caption{Calibration deviation (D21) $\downarrow$ stratified across IRSD quintiles.}\label{tab:4Cali}
\begin{tabular}{lccccc}
\toprule
\textbf{Stratum} & \textbf{Real} & \textbf{WGAN-GP} & \textbf{WGAN-GP+VAE} & \textbf{DDPM} & \textbf{MCM} \\
\midrule

Q1 (Most disadvantaged)
& 0.05 & 0.21 & 0.91 & 0.61 & 0.35 \\

Q2
& 0.06 & 0.15 & 0.58 & 0.80 & 0.43 \\

Q3
& 0.04 & 0.19 & 0.81 & 0.64 & 0.45 \\

Q4
& 0.08 & 0.21 & 0.73 & 0.22 & 0.04 \\

Q5 (Least disadvantaged)
& 0.06 & 0.35 & 0.99 & 0.80 & 0.31 \\

\bottomrule
\end{tabular}

\end{table}

but lower precision (0.29--0.41), consistent with overestimation of edge presence in graphs. For orientation, performance is uniformly lower across all models. MCM again achieves the highest F1 score (0.52 [0.35--0.63]), whereas other models range from 0.18 to 0.42. 

\paragraph{Supplementary details.}
Numerical results underlying all figures are reported in Appendices \S~\ref{App:MoreResultsStream1}--\ref{App:MoreResultsStream2}; we also provide additional analyses on the stratification by age groups.

\section{Discussion}\label{Sec:Dis}

\paragraph{Evaluation summary.} No model simultaneously achieves strong performance in descriptive fidelity, clinical utility, and structural validity. At the population level, all models reproduce marginal distributions (Figure~\ref{fig:4Dist}); however, this does not extend to clinically meaningful subgroups (Figure~\ref{fig:4DistStratIRSD}).

Reduced subgroup fidelity is not consistently reflected in covariate effect estimates. As shown in Figure~\ref{fig:4ClinicalUtility}(a), HRs derived from synthetic data of WGAN-GP, WGAN-GP+VAE, and MCM are broadly aligned with the ground truth; suggesting relative effects can be preserved despite subgroup distortion.

\newpage
\begin{figure}[h!]
    \centering
    \includegraphics[width=\linewidth]{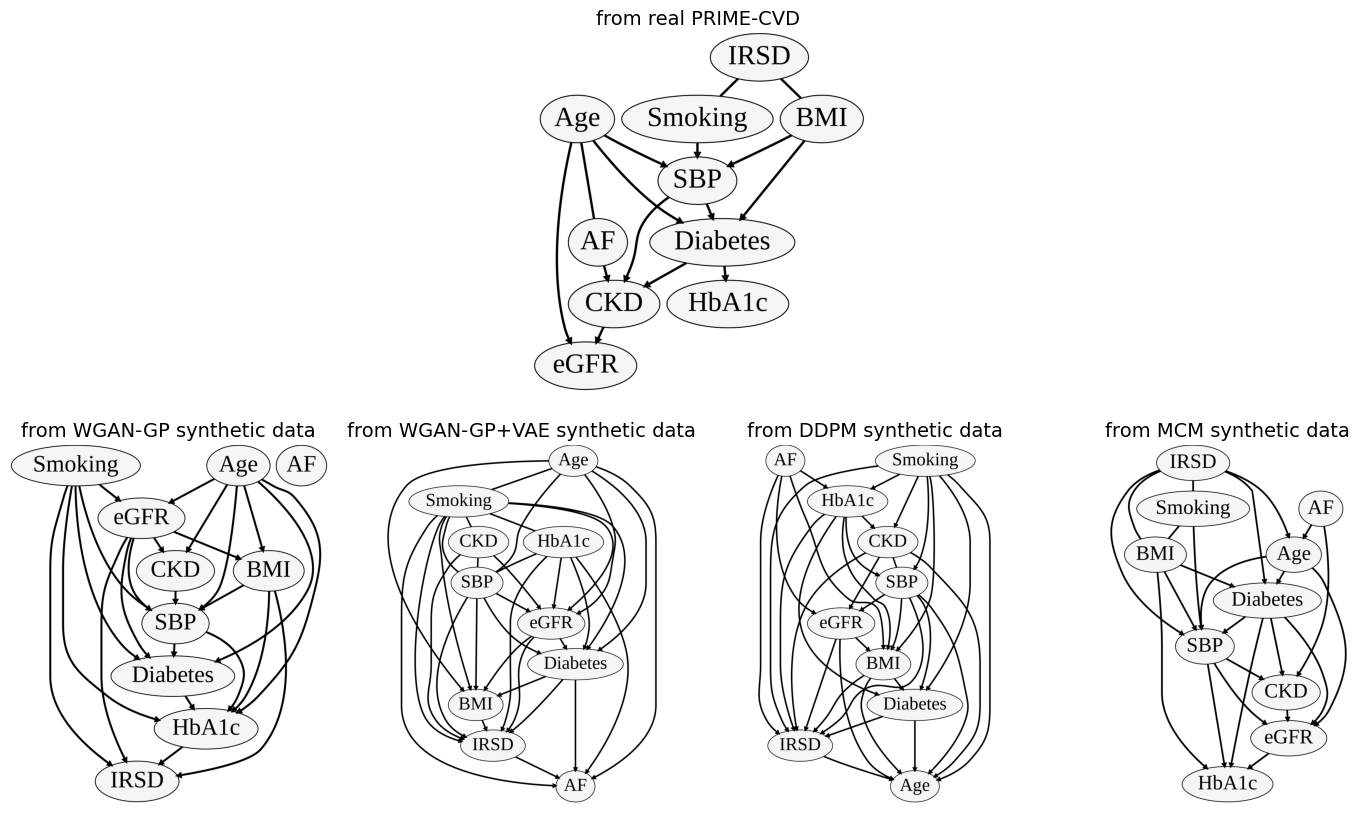}
    \caption{Structural validity comparison using DAGs discovered through GES}
    \label{fig:4DAGs}
\end{figure}

\begin{table}[h!]
\centering
\caption{Structural validity comparison reported in median [interquartile range].}\label{Tab:4DAGsStability}
\begin{tabular}{lllll}
\toprule
\textbf{Metric} & \textbf{WGAN-GP} & \textbf{WGAN-GP+VAE} & \textbf{DDPM} & \textbf{MCM} \\
\midrule

\multicolumn{5}{l}{\textbf{Structural Distance}} \\

SHD $\downarrow$
& 18 [14--21]
& 26 [23--28]
& 32 [25--35]
& 16 [11--21] \\

\midrule
\multicolumn{5}{l}{\textbf{Adjacency Metrics}} \\

Precision $\uparrow$
& 0.41 [0.36--0.48]
& 0.30 [0.27--0.38]
& 0.29 [0.24--0.35]
& 0.46 [0.37--0.55] \\

Recall $\uparrow$
& 0.83 [0.67--0.83]
& 0.58 [0.58--0.92]
& 0.79 [0.58--0.92]
& 0.92 [0.67--1.00] \\

F1 Score $\uparrow$
& 0.54 [0.49--0.61]
& 0.40 [0.37--0.54]
& 0.43 [0.34--0.51]
& 0.62 [0.48--0.69] \\

\midrule
\multicolumn{5}{l}{\textbf{Orientation Metrics}} \\

Precision $\uparrow$
& 0.32 [0.27--0.43]
& 0.16 [0.15--0.23]
& 0.12 [0.03--0.21]
& 0.38 [0.27--0.50] \\

Recall $\uparrow$
& 0.67 [0.50--0.75]
& 0.33 [0.33--0.58]
& 0.33 [0.08--0.58]
& 0.83 [0.50--0.92] \\

F1 Score $\uparrow$
& 0.42 [0.37--0.55]
& 0.22 [0.21--0.33]
& 0.18 [0.05--0.30]
& 0.52 [0.35--0.63] \\

\bottomrule
\end{tabular}
\end{table}

Calibration, however, is more sensitive to these discrepancies. Although WGAN-GP's synthetic data performs well at the population level (Figure~\ref{fig:4ClinicalUtility}(b)), all synthetic data exhibit increased error under stratification (Table~\ref{tab:4Cali}), indicating inconsistent risk estimation across subpopulations.

Structural evaluation further differentiates models. Synthetic data-derived graphs are generally denser than the ground truth (Figure~\ref{fig:4DAGs}), reflected in lower adjacency precision and indicating spurious relationships, with even poorer recovery of edge directionality (Table~\ref{Tab:4DAGsStability}).

\paragraph{Concluding remarks.}
These findings have important implications for synthetic data use in healthcare. From an epidemiological perspective, the three evaluation streams map to descriptive, predictive, and causal questions~\cite{carlin2026identifying, dyer2025distinction}; current models do not reliably support all three simultaneously. While effect estimates may appear preserved, degraded calibration and subgroup inconsistency limit predictive reliability, and spurious dependencies constrain causal use. This may explain why synthetic data can outperform SMOTE~\cite{chawla2002smote} in subgroups augmentation, yet yield unstable gains~\cite{kuo2024ck4gen}.

While privacy remain important, our results challenge the realism-privacy trade-off~\cite{wang2021generating, dwork2025differential}, suggesting fidelity remains insufficient for such concerns to dominate. More critically, incorporating unrealistic synthetic data into decision models risks introducing bias and may ultimately harm real patients. We acknowledge that our evaluation is conducted on a single dataset PRIME-CVD with known ground-truth structure, which may not capture all complexities of real-world EMR data.

\newpage
\bibliographystyle{unsrtnat}
\bibliography{A001_References}

\appendix
\newpage
\section*{Appendix: Additional Details to the Main Text}

\paragraph{Purpose of this Appendix.}
This appendix provides detailed methodological, implementation, and supplementary result information supporting the main text. It ensures reproducibility, clarifies evaluation procedures, and presents extended analyses that substantiate the empirical findings across our evaluation streams.

\paragraph{Contents.}
\begin{itemize}

    \item \textbf{Reproducibility}
    \begin{itemize}
        \item \S~\ref{App:Repro_Dependencies} Code Dependencies \dotfill \pageref{App:Repro_Dependencies}
        \item \S~\ref{App:Repro_CodeAvail} Model Availability \dotfill \pageref{App:Repro_CodeAvail}
    \end{itemize}

    \item \textbf{Model Details}
    \begin{itemize}
        \item \S~\ref{App:Model1_All} WGAN-GP \dotfill \pageref{App:Model1_All}
        \item \hspace{5mm} Hyperparameters \dotfill \pageref{App:Model1_Hyperparameters}
        \item \S~\ref{App:Model2_All} WGAN-GP + VAE \dotfill \pageref{App:Model2_All}
        \item \hspace{5mm} Hyperparameters \dotfill \pageref{App:Model2_Hyperparameters}
        \item \S~\ref{App:Model3_All} Diffusion (DDPM) \dotfill \pageref{App:Model3_All}
        \item \hspace{5mm} Hyperparameters \dotfill \pageref{App:Model3_Hyperparameters}
        \item \S~\ref{App:Model4_All} Masked Clinical Model (MCM) \dotfill \pageref{App:Model4_All}
        \item \hspace{5mm} Hyperparameters \dotfill \pageref{App:Model4_Hyperparameters}
    \end{itemize}

    \item \textbf{Framework Details}
    \begin{itemize}
        \item \S~\ref{App:InfStream1} Descriptive Fidelity (Stream 1) \dotfill \pageref{App:InfStream1}
        \item \S~\ref{App:InfStream2}  Clinical Utility (Stream 2) \dotfill \pageref{App:InfStream2}
        \item \hspace{5mm} \S~\ref{App:InfStream2_IntroduceCox}  Cox Proportional Hazards \dotfill \pageref{App:InfStream2_IntroduceCox}
        \item \hspace{5mm} \S~\ref{App:InfStream2_IntroduceMetrics} Calibration Metrics (D21) \dotfill \pageref{App:InfStream2_IntroduceMetrics}
        \item \S~\ref{App:InfStream3} Structural Validity (Stream 3) \dotfill \pageref{App:InfStream3}
        \item \hspace{5mm} \S~\ref{App:InfStream3_IntroduceGES}  GES Details \dotfill \pageref{App:InfStream3_IntroduceGES}
        \item \hspace{5mm} \S~\ref{App:InfStream3_StrucMetrics} Structural Metrics \dotfill \pageref{App:InfStream3_StrucMetrics}
    \end{itemize}

    \item \textbf{Additional Results}
    \begin{itemize}
        \item \S~\ref{App:MoreResultsStream1} Descriptive Fidelity Results \dotfill \pageref{App:MoreResultsStream1}
        \item \hspace{5mm} Age-Stratified Analysis \dotfill \pageref{sec:appendix_age_strat}
        \item \S~\ref{App:MoreResultsStream2} Clinical Utility Results \dotfill \pageref{App:MoreResultsStream2}
    \end{itemize}    
\end{itemize}

\paragraph{Reproducibility.}
Specifies software, hardware, and deterministic settings, along with model availability, ensuring all experiments can be reliably reproduced.

\paragraph{Model Details.}
Provides full architectural, objective, and optimisation details for all generative models (WGAN-GP, Hybrid, DDPM, MCM), including hyperparameters. Related codes are submitted as supplementary text files to OpenReview.

\paragraph{Framework Details.}
Defines the three-stream evaluation framework (descriptive fidelity, clinical utility, and structural validity), including formal procedures, model formulations, and evaluation metrics.

\paragraph{Additional Results.}
Presents extended quantitative results supporting the main text, including subgroup analyses and detailed effect estimates.

\newpage
\section{Reproducibility and Implementation Details}\label{App:Repro}
\subsection{Code Dependencies}\label{App:Repro_Dependencies}
We provide a detailed account of the software and hardware dependencies required to reproduce our experiments. All experiments were conducted in a controlled environment with deterministic computation enabled, ensuring consistent results across runs.

All code was executed using Python~3.12.13~\cite{van1995python}. The implementation relies on a standard scientific computing stack, complemented by specialised libraries for deep learning, survival analysis, and causal discovery.

Our implementation is based on PyTorch with GPU acceleration. The framework is compiled against CUDA~12.8, with cuDNN support enabled. Experiments were conducted on a single NVIDIA Tesla T4 GPU to provide sufficient memory for all models evaluated in this work.

To ensure reproducibility, we fix all random seeds across Python, NumPy, and PyTorch. Furthermore, we enforce deterministic behaviour for GPU operations where supported.

\begin{table}[h]
\centering
\caption{Software, hardware, and reproducibility configuration.}
\begin{tabular}{lll}
\toprule
\textbf{Category} & \textbf{Component} & \textbf{Specification} \\
\midrule

\multicolumn{3}{l}{\textit{Software Environment}} \\
 & Python~\cite{van1995python} & 3.12.13 \\
 & NumPy~\cite{harris2020array} & 2.0.2 \\
 & Pandas~\cite{mckinney2010data} & 2.2.2 \\
 & SciPy~\cite{virtanen2020scipy} & 1.16.3 \\
 & scikit-learn~\cite{pedregosa2011scikit} & 1.6.1 \\
 & Matplotlib~\cite{hunter2007matplotlib} & 3.10.0 \\
 & Seaborn~\cite{waskom2021seaborn} & 0.13.2 \\
 & Lifelines~\cite{davidson2019lifelines} & 0.30.3 \\
 & PyTorch~\cite{paszke2019pytorch} & 2.10.0+cu128 \\
 & Causal-learn~\cite{zheng2024causal} & 0.1.4.5 \\

\midrule
\multicolumn{3}{l}{\textit{Deep Learning and CUDA}} \\
 & CUDA version (PyTorch build) & 12.8 \\
 & cuDNN version & 91002 \\

\midrule
\multicolumn{3}{l}{\textit{Hardware}} \\
 & GPU model & NVIDIA Tesla T4 \\
 & GPU memory & 16 GB \\

\midrule
\multicolumn{3}{l}{\textit{Determinism}} \\
 & Global random seed & 42 \\
 & Deterministic algorithms & Enabled \\
 & cuDNN deterministic mode & Enabled\\

\bottomrule
\end{tabular}
\end{table}

\subsection{Model Availabilities}\label{App:Repro_CodeAvail}

All models required to reproduce our results are provided as four supplementary text files on OpenReview, corresponding to WGAN-GP, WGAN-GP+VAE, DDPM, and MCM. Each file contains the complete implementation, including architecture, forward pass, loss functions, and training procedure exactly as used in our experiments. Combined with the formal descriptions in the following appendix sections, and the explicitly specified software and hardware configuration, these materials are sufficient for AI/ML practitioners to replicate the results reported in this paper.

The entire pipeline will be made publicly available in a dedicated GitHub repository after paper acceptance.

\newpage
\section{Additional Details on Model 1: WGAN-GP}\label{App:Model1_All}
We describe a Wasserstein Generative Adversarial Network with Gradient Penalty (WGAN-GP)~\cite{kuo2022health} adapted for static tabular healthcare data with mixed variable types. The model is designed to generate samples that preserve both marginal distributions and inter-variable dependencies.

\subsection{Problem Setup}

Let $X_{\mathrm{real}} \sim P_{\mathrm{data}}$ denote real tabular data, where each observation is a vector:
\[
x \in \mathbb{R}^p.
\]

Let $z \sim P_z$ denote latent noise, where:
\[
z \sim \mathcal{U}(0,1)^d.
\]

The generator $G$ induces a distribution $P_G$ over the data space:
\[
x_{\mathrm{syn}} = G(z), \quad P_G \approx P_{\mathrm{data}}.
\]

\subsection{Symbol Glossary}

\begin{table}[h]
\centering
\begin{tabular}{ll}
\toprule
Symbol & Meaning \\
\midrule
$x$ & Data vector (tabular observation) \\
$z$ & Latent noise vector \\
$G$ & Generator network \\
$D$ & Critic (discriminator) network \\
$p$ & Total feature dimension (after encoding) \\
$d$ & Latent dimension \\
$\lambda_{\mathrm{GP}}$ & Gradient penalty weight \\
$\lambda_{\mathrm{corr}}$ & Correlation loss weight \\
$r_{ij}$ & Pearson correlation between variables $i,j$ \\
\bottomrule
\end{tabular}
\caption{Notation used in the static generative model.}
\end{table}

\subsection{Data Representation}

The data consist of mixed variable types:
\begin{itemize}
\item Continuous variables, transformed to $[0,1]$ via Box-Cox and min-max scaling,
\item Binary and categorical variables represented as one-hot vectors.
\end{itemize}

Let the feature vector be partitioned as:
\[
x = \big(x_{\mathrm{real}}, x^{(1)}_{\mathrm{cat}}, \dots, x^{(K)}_{\mathrm{cat}}\big),
\]
where $x_{\mathrm{real}} \in [0,1]^{p_r}$ and each $x^{(k)}_{\mathrm{cat}} \in \Delta^{C_k-1}$ lies on a probability simplex.

\subsection{Generator}

The generator is a fully connected neural network:
\[
G: \mathbb{R}^d \rightarrow \mathbb{R}^p,
\quad x = G(z) = f_\theta(z),
\]
where $f_\theta$ is a multi-layer perceptron with nonlinear activations.

The output is structured as:
\[
x_{\mathrm{real}} = \sigma(g_{\mathrm{real}}(z)), \quad
x^{(k)}_{\mathrm{cat}} = \mathrm{softmax}(g^{(k)}(z)),
\]
ensuring:
\begin{itemize}
\item continuous variables lie in $[0,1]$,
\item categorical variables form valid probability distributions.
\end{itemize}

\subsection{Critic}

The critic maps input vectors to scalar realism scores:
\[
D: \mathbb{R}^p \rightarrow \mathbb{R}.
\]

To handle categorical variables, we employ soft embeddings. For each categorical block $x^{(k)}$, define:
\[
e^{(k)} = x^{(k)} W^{(k)},
\]
where $W^{(k)} \in \mathbb{R}^{C_k \times d_k}$ is an embedding matrix.

The critic input is:
\[
h = \big(x_{\mathrm{real}}, e^{(1)}, \dots, e^{(K)}\big),
\]
and:
\[
D(x) = f_\phi(h),
\]
where $f_\phi$ is a feedforward network.

\subsection{Wasserstein Objective}

The critic is trained to minimise:
\[
\mathcal{L}_D =
\mathbb{E}_{x_{\mathrm{syn}}}[D(x_{\mathrm{syn}})]
- \mathbb{E}_{x_{\mathrm{real}}}[D(x_{\mathrm{real}})]
+ \lambda_{\mathrm{GP}} \, \mathbb{E}_{\hat{x}}
\left[
\left(\|\nabla_{\hat{x}} D(\hat{x})\|_2 - 1\right)^2
\right],
\]
where:
\[
\hat{x} = \epsilon x_{\mathrm{real}} + (1-\epsilon)x_{\mathrm{syn}}, \quad \epsilon \sim \mathcal{U}(0,1).
\]

The generator is trained to minimise:
\[
\mathcal{L}_G^{\mathrm{WGAN}} =
- \mathbb{E}_{z}[D(G(z))].
\]

\subsection{Correlation Alignment Loss}

To preserve dependency structure, we compute correlation matrices:
\[
R^{\mathrm{real}}, \quad R^{\mathrm{syn}}.
\]

A stochastic regularisation is applied:
\[
\tilde{R}_{ij} = \min(u_{ij}, |R_{ij}|)\cdot \mathrm{sign}(R_{ij}), \quad u_{ij} \sim \mathcal{U}(0,1).
\]

The alignment loss is:
\[
\mathcal{L}_{\mathrm{corr}} =
\|\tilde{R}^{\mathrm{syn}} - \tilde{R}^{\mathrm{real}}\|_1.
\]

The final generator objective is:
\[
\mathcal{L}_G =
- \mathbb{E}_{z}[D(G(z))]
+ \lambda_{\mathrm{corr}} \, \mathcal{L}_{\mathrm{corr}}.
\]

\subsection{Optimisation}

The model is trained via alternating updates:
\[
\min_G \max_D \;
\mathbb{E}_{x_{\mathrm{real}}}[D(x_{\mathrm{real}})]
- \mathbb{E}_{z}[D(G(z))]
+ \lambda_{\mathrm{GP}} \cdot \mathrm{GP}
+ \lambda_{\mathrm{corr}} \cdot \mathcal{L}_{\mathrm{corr}}.
\]

\newpage
\subsection{Hyperparameters}\label{App:Model1_Hyperparameters}
\begin{table}[h]
\centering
\small
\begin{tabular}{lll}
\toprule
\textbf{Category} & \textbf{Hyperparameter} & \textbf{Value} \\
\midrule

\multirow{3}{*}{Training} 
& Batch size & 256 \\
& Epochs & 20 \\
& Critic updates per step ($k$) & 5 \\

\midrule
\multirow{3}{*}{Optimisation} 
& Optimiser & Adam~\cite{kingma2014adam} \\
& Learning rate & $1 \times 10^{-3}$ \\
& Adam $(\beta_1, \beta_2)$ & $(0.9, 0.99)$ \\

\midrule
\multirow{2}{*}{WGAN-GP} 
& Gradient penalty ($\lambda_{\mathrm{GP}}$) & 10.0 \\
& Correlation loss ($\lambda_{\mathrm{corr}}$) & 10.0 \\

\midrule
\multirow{4}{*}{Architecture} 
& Latent dimension & 128 \\
& Hidden dimension & 128 \\
& Number of hidden layers & 3 \\
& Activation & LeakyReLU (0.1) \\

\midrule
\multirow{4}{*}{Data Representation} 
& Continuous transform & Box-Cox + MinMax $[0,1]$ \\
& Latent prior & $\mathcal{U}(0,1)$ \\
& Categorical encoding & One-hot (softmax output) \\
& Binary encoding & 2-dim one-hot \\

\midrule
\multirow{2}{*}{Embeddings} 
& Binary embedding size & 2 \\
& Categorical embedding size & 4 \\

\midrule
\multirow{3}{*}{Correlation Loss} 
& Loss type & L1 (mean) \\
& Correlation estimate & Batch-wise Pearson \\
& Regularisation & Random clipping ($\mathcal{U}(0,1)$) \\

\bottomrule
\end{tabular}
\caption{Hyperparameters used for training the WGAN-GP model.}
\label{tab:wgan_hyperparameters}
\end{table}

\newpage
\section{Additional Details on Model 2: WGAN-GP+VAE}\label{App:Model2_All}
We describe a hybrid WGAN-GP for static tabular healthcare data with mixed variable types. The model combines a schema-aware transformer-based generator, a soft-embedding critic, and an auxiliary VAE used to construct a data-informed latent sampling mechanism~\cite{nicholas2023generating}. The aim is to generate synthetic samples that preserve marginal distributions, mixed-type structure, and inter-variable dependencies.

\subsection{Problem Setup}

Let $X_{\mathrm{real}} \sim P_{\mathrm{data}}$ denote real tabular data, where each observation is represented as
\[
x \in \mathbb{R}^p.
\]

Let $z \sim P_z$ denote a latent vector in a $d$-dimensional latent space:
\[
z \in \mathbb{R}^d.
\]

The generator $G$ induces a synthetic distribution $P_G$ over the encoded data space:
\[
x_{\mathrm{syn}} = G(z), \quad P_G \approx P_{\mathrm{data}}.
\]

Unlike a standard GAN with a fixed latent prior, the present model employs an auxiliary latent model to construct an empirical, data-adaptive sampling distribution for $z$.

\subsection{Symbol Glossary}

\begin{table}[h]
\centering
\begin{tabular}{ll}
\toprule
Symbol & Meaning \\
\midrule
$x$ & Encoded tabular observation \\
$z$ & Latent vector \\
$G$ & Generator network \\
$D$ & Critic network \\
$A$ & Auxiliary variational autoencoder \\
$p$ & Total encoded feature dimension \\
$d$ & Latent dimension \\
$h_j$ & Token representation for feature $j$ \\
$e^{(k)}$ & Soft embedding for categorical block $k$ \\
$\lambda_{\mathrm{GP}}$ & Gradient penalty weight \\
$\lambda_{\mathrm{corr}}$ & Correlation alignment weight \\
$\mu$ & Latent mean estimated by the auxiliary encoder \\
$\sigma$ & Latent standard deviation used for latent perturbation \\
$R^{\mathrm{real}}$ & Correlation matrix of real encoded data \\
$R^{\mathrm{syn}}$ & Correlation matrix of synthetic encoded data \\
\bottomrule
\end{tabular}
\caption{Notation used in the hybrid static generative model.}
\end{table}

\subsection{Generator}

The generator is a schema-aware transformer-based network:
\[
G:\mathbb{R}^d \rightarrow \mathbb{R}^p.
\]

Given a latent vector $z$, the generator first projects it into a collection of feature tokens:
\[
H^{(0)} = \mathrm{reshape}(W_{\mathrm{in}} z + b_{\mathrm{in}})
\in \mathbb{R}^{T \times h},
\]
where $T$ is the number of schema-defined variables and $h$ is the hidden dimension.

These tokens are then passed through a transformer encoder:
\[
H = \mathrm{TransformerEncoder}\!\left(H^{(0)}\right),
\]
allowing interactions across variable-specific token representations.

A pooled global representation is obtained by averaging across tokens:
\[
\bar{h} = \frac{1}{T}\sum_{j=1}^T H_j.
\]

Finally, separate schema-specific output heads decode $\bar{h}$ into the output blocks:
\[
x_{\mathrm{real}}^{(j)} = \sigma(g_j(\bar{h})),
\qquad
x_{\mathrm{cat}}^{(k)} = \mathrm{softmax}(g_k(\bar{h})),
\qquad
x_{\mathrm{bin}}^{(b)} = \mathrm{softmax}(g_b(\bar{h})).
\]

Thus:
\begin{itemize}
\item continuous variables are constrained to $[0,1]$,
\item binary and categorical outputs form valid probability vectors,
\item each variable is decoded by its own schema-aware head.
\end{itemize}

This design allows the generator to model cross-feature dependence through transformer interactions while preserving variable-type constraints at the output layer.

\subsection{Auxiliary Latent Autoencoder}

In addition to the adversarial generator--critic pair, the model contains an auxiliary variational autoencoder $A$ trained on the critic's embedded representation of real data. Let
\[
h_{\mathrm{emb}} = E_D(x)
\]
denote the embedded critic representation prior to the critic's final scoring layers.

The auxiliary encoder maps this representation to latent parameters:
\[
(\mu, \log \sigma^2) = A_{\mathrm{enc}}(h_{\mathrm{emb}}).
\]

A latent code is sampled using the reparameterisation trick:
\[
z = \mu + \epsilon \odot \sigma,
\qquad
\epsilon \sim \mathcal{N}(0, I).
\]

The auxiliary decoder reconstructs the critic-embedded input:
\[
\hat{h}_{\mathrm{emb}} = A_{\mathrm{dec}}(z).
\]

The auxiliary loss is
\[
\mathcal{L}_{A}
=
\| \hat{h}_{\mathrm{emb}} - h_{\mathrm{emb}} \|_2^2
+
\mathcal{L}_{\mathrm{KL}},
\]
where
\[
\mathcal{L}_{\mathrm{KL}}
=
-\frac{1}{2}\,
\mathbb{E}\!\left[
1 + \log \sigma^2 - \mu^2 - \sigma^2
\right].
\]

During training, the model stores a buffer of latent means together with a running latent standard deviation estimate. These are later used to define a data-adaptive latent sampling distribution for the generator.

\subsection{Latent Sampling Mechanism}

Instead of always drawing latent vectors from a fixed standard Gaussian prior, the model samples from an empirical latent mixture induced by the auxiliary autoencoder.

If the latent buffer has been sufficiently populated, a mean vector $\mu_i$ is sampled from the stored latent means and perturbed as
\[
z = \mu_i + \epsilon \odot \sigma_{\mathrm{buffer}},
\qquad
\epsilon \sim \mathcal{N}(0, I),
\]
where $\sigma_{\mathrm{buffer}}$ is the running latent standard deviation estimate.

If the latent buffer is not yet populated, latent vectors are sampled from
\[
z \sim \mathcal{N}(0, I).
\]

This mechanism biases generation toward regions of latent space supported by real data, while retaining stochastic variation through Gaussian perturbation.

\subsection{Optimisation}

Training proceeds by alternating three updates within each minibatch:
\begin{enumerate}
\item an auxiliary autoencoder update on embedded real samples,
\item multiple critic updates using the WGAN-GP objective,
\item one generator update using the adversarial and correlation-alignment losses.
\end{enumerate}

Formally, the model optimises
\[
\min_{G,A}\max_D
\;
\mathbb{E}_{x_{\mathrm{real}}}[D(x_{\mathrm{real}})]
-
\mathbb{E}_{z}[D(G(z))]
+
\lambda_{\mathrm{GP}} \cdot \mathrm{GP}
+
\lambda_{\mathrm{corr}} \cdot \mathcal{L}_{\mathrm{corr}}
+
\mathcal{L}_{A},
\]
with the understanding that $\mathcal{L}_A$ is applied only to the auxiliary autoencoder parameters, while the adversarial objectives govern the generator and critic updates.

In the implementation, Adam optimisation is used for all three components, and each minibatch applies one auxiliary update, five critic updates, and one generator update.

\subsection{Hyperparameters}\label{App:Model2_Hyperparameters}
\begin{table}[h]
\centering
\small
\begin{tabular}{lll}
\toprule
\textbf{Category} & \textbf{Hyperparameter} & \textbf{Value} \\
\midrule

\multirow{3}{*}{Training} 
& Batch size & 256 \\
& Epochs & 5 \\
& Critic updates per step ($k$) & 5 \\

\midrule
\multirow{3}{*}{Optimisation} 
& Optimiser & Adam \\
& Learning rate & $1 \times 10^{-3}$ \\
& Adam $(\beta_1, \beta_2)$ & $(0.9, 0.99)$ \\

\midrule
\multirow{2}{*}{WGAN-GP} 
& Gradient penalty ($\lambda_{\mathrm{GP}}$) & 10.0 \\
& Correlation loss ($\lambda_{\mathrm{corr}}$) & 10.0 \\

\midrule
\multirow{6}{*}{Architecture} 
& Latent dimension & 128 \\
& Hidden dimension & 128 \\
& Transformer / MLP layers & 1 \\
& Attention heads & 4 \\
& Feedforward dimension & 256 \\
& Activation & GELU + LeakyReLU (0.1) \\

\midrule
\multirow{6}{*}{VAE (Auxiliary)} 
& Latent dimension & 128 \\
& Hidden dimension & 128 \\
& Encoder layers & 2 \\
& Decoder layers & 2 \\
& Reconstruction loss & MSE \\
& KL divergence & Standard VAE KL \\

\midrule
\multirow{4}{*}{Latent Sampling} 
& Base prior & $\mathcal{N}(0, I)$ \\
& Sampling rule & $\mu + \epsilon \cdot \sigma$ \\
& Latent buffer size (declared) & 500,000 \\
& Latent buffer size (effective) & 10,000 \\

\midrule
\multirow{4}{*}{Data Representation} 
& Continuous transform & Box-Cox + MinMax $[0,1]$ \\
& Box-Cox $\epsilon$ & $10^{-3}$ \\
& Categorical encoding & One-hot (softmax output) \\
& Binary encoding & 2-dim one-hot \\

\midrule
\multirow{3}{*}{Embeddings} 
& Real embedding size & 1 \\
& Binary embedding size & 2 \\
& Categorical embedding size & 4 \\

\midrule
\multirow{3}{*}{Correlation Loss} 
& Loss type & L1 \\
& Correlation estimate & Batch-wise Pearson \\
& Numerical stability ($\epsilon$) & $10^{-8}$ \\

\bottomrule
\end{tabular}
\caption{Hyperparameters used for training the hybrid WGAN-GP model with VAE sampling.}
\label{tab:hybrid_wgan_hyperparameters}
\end{table}

\newpage
\section{Additional Details on Model 3: DDPM}\label{App:Model3_All}

We describe a DDPM~\cite{nicholas2023synthetic} adapted for static tabular healthcare data with mixed variable types. The model replaces adversarial training with a noise-driven generative process and incorporates schema-aware constraints and correlation regularisation to preserve marginal distributions and inter-variable dependencies.

\subsection{Problem Setup}

Let $X_{\mathrm{real}} \sim P_{\mathrm{data}}$ denote real tabular data, where each observation is represented as
\[
x_0 \in \mathbb{R}^p.
\]

A forward diffusion process progressively corrupts $x_0$ into noisy versions $x_t$:
\[
x_t \sim q(x_t \mid x_0),
\quad t = 1,\dots,T.
\]

The model learns a parameterised reverse process:
\[
p_\theta(x_{t-1} \mid x_t),
\]
such that sampling from Gaussian noise $x_T \sim \mathcal{N}(0, I)$ and iteratively applying the reverse transitions yields synthetic samples:
\[
x_{\mathrm{syn}} \approx x_0.
\]

\subsection{Symbol Glossary}

\begin{table}[h]
\centering
\begin{tabular}{ll}
\toprule
Symbol & Meaning \\
\midrule
$x_0$ & Clean encoded tabular observation \\
$x_t$ & Noisy observation at timestep $t$ \\
$\epsilon$ & Gaussian noise \\
$\hat{\epsilon}$ & Predicted noise \\
$T$ & Total number of diffusion steps \\
$\beta_t$ & Noise variance schedule \\
$\alpha_t$ & Noise retention coefficient \\
$\bar{\alpha}_t$ & Cumulative product of $\alpha_t$ \\
$f_\theta$ & Denoising neural network \\
$p$ & Total encoded feature dimension \\
$R^{\mathrm{real}}$ & Correlation matrix of real data \\
$R^{\mathrm{syn}}$ & Correlation matrix of synthetic data \\
\bottomrule
\end{tabular}
\caption{Notation used in the static diffusion model.}
\end{table}

\subsection{Data Representation}

The data consist of mixed variable types:
\begin{itemize}
\item Continuous variables, transformed by Box--Cox transformation followed by min--max scaling to $[0,1]$,
\item Binary variables represented as two-dimensional one-hot blocks,
\item Multi-class categorical variables represented as one-hot vectors.
\end{itemize}

Accordingly, the encoded feature vector is partitioned as
\[
x = \big(x_{\mathrm{real}}, x_{\mathrm{bin}}^{(1)}, \dots, x_{\mathrm{bin}}^{(B)}, x_{\mathrm{cat}}^{(1)}, \dots, x_{\mathrm{cat}}^{(K)}\big),
\]
where
\[
x_{\mathrm{real}} \in [0,1]^{p_r},
\]
and each discrete block lies on a simplex:
\[
x_{\mathrm{bin}}^{(b)} \in \Delta^{1}, \qquad
x_{\mathrm{cat}}^{(k)} \in \Delta^{C_k-1}.
\]

A schema object defines index spans and types for each variable, enabling structured processing and valid decoding.

\subsection{Denoising Network}

The model uses a timestep-conditioned neural network:
\[
f_\theta(x_t, t): \mathbb{R}^p \times \{1,\dots,T\} \rightarrow \mathbb{R}^p,
\]
which predicts the noise component added at timestep $t$.

The timestep $t$ is embedded via a learned embedding:
\[
e_t = \mathrm{Embedding}(t),
\]
and concatenated with the noisy input:
\[
h = \big(x_t, e_t\big).
\]

The network consists of a multi-layer perceptron with LayerNorm and SiLU nonlinearities:
\[
h^{(l+1)} = \phi\big(\mathrm{LayerNorm}(W_l h^{(l)} + b_l)\big),
\]
followed by a linear projection to the original feature dimension.

\subsection{Schema Projection}

To ensure valid tabular outputs, the reconstructed sample $\hat{x}_0$ is projected back to the schema:
\[
x_{\mathrm{real}} = \sigma(\hat{x}_0^{\mathrm{real}}), \qquad
x_{\mathrm{cat}}^{(k)} = \mathrm{softmax}(\hat{x}_0^{(k)}).
\]

This guarantees:
\begin{itemize}
\item continuous variables remain in $[0,1]$,
\item categorical and binary variables form valid probability distributions.
\end{itemize}

\subsection{Forward Diffusion Process}

The forward process corrupts clean data by adding Gaussian noise:
\[
x_t = \sqrt{\bar{\alpha}_t} \, x_0 + \sqrt{1 - \bar{\alpha}_t} \, \epsilon,
\quad \epsilon \sim \mathcal{N}(0, I).
\]

The variance schedule is defined by:
\[
\beta_t \in [\beta_{\min}, \beta_{\max}],
\]
with linearly spaced values across $T$ steps.

\subsection{Reverse Process}

The model learns to predict noise:
\[
\hat{\epsilon} = f_\theta(x_t, t),
\]
and reconstruct the clean sample:
\[
\hat{x}_0 =
\frac{x_t - \sqrt{1 - \bar{\alpha}_t} \, \hat{\epsilon}}
{\sqrt{\bar{\alpha}_t}}.
\]

Synthetic samples are generated by iteratively applying:
\[
x_{t-1} = \frac{1}{\sqrt{\alpha_t}}\left(
x_t - \frac{1 - \alpha_t}{\sqrt{1 - \bar{\alpha}_t}} \hat{\epsilon}
\right) + \sqrt{\beta_t} \, \eta,
\quad \eta \sim \mathcal{N}(0, I),
\]
starting from $x_T \sim \mathcal{N}(0, I)$.

\newpage
\subsection{Training Objective}

The model is trained using a composite objective:
\[
\mathcal{L}
=
\mathcal{L}_{\mathrm{diff}}
+
\lambda_{\mathrm{rec}} \, \mathcal{L}_{\mathrm{rec}}
+
\lambda_{\mathrm{corr}} \, \mathcal{L}_{\mathrm{corr}}.
\]

Diffusion loss:
\[
\mathcal{L}_{\mathrm{diff}} =
\|\hat{\epsilon} - \epsilon\|_2^2.
\]

Reconstruction loss:
\[
\mathcal{L}_{\mathrm{rec}} =
\|\hat{x}_0 - x_0\|_2^2.
\]

Correlation alignment loss:
\[
\mathcal{L}_{\mathrm{corr}} =
\|R^{\mathrm{syn}} - R^{\mathrm{real}}\|_1.
\]

This encourages preservation of global dependency structure in addition to denoising accuracy.

\subsection{Optimisation}

Training proceeds by sampling random timesteps $t$ and optimising the denoising objective over minibatches:
\begin{enumerate}
\item sample $x_0$ from real data,
\item sample timestep $t \sim \mathcal{U}\{1,\dots,T\}$,
\item generate $x_t$ via forward diffusion,
\item predict noise $\hat{\epsilon} = f_\theta(x_t, t)$,
\item compute loss and update parameters.
\end{enumerate}

The model is optimised using Adam, and generation is performed via iterative reverse diffusion over $T$ steps.

\newpage
\subsection{Hyperparameters}\label{App:Model3_Hyperparameters}
\begin{table}[h]
\centering
\small
\begin{tabular}{lll}
\toprule
\textbf{Category} & \textbf{Hyperparameter} & \textbf{Value} \\
\midrule

\multirow{3}{*}{Training} 
& Batch size & 256 \\
& Epochs & 20 \\

\midrule
\multirow{3}{*}{Optimisation} 
& Optimiser & Adam \\
& Learning rate & $1 \times 10^{-3}$ \\
& Adam $(\beta_1, \beta_2)$ & $(0.9, 0.99)$ \\

\midrule
\multirow{3}{*}{Diffusion} 
& Diffusion steps ($T$) & 200 \\
& $\beta_{\min}$ & $1 \times 10^{-4}$ \\
& $\beta_{\max}$ & $1 \times 10^{-2}$ \\

\midrule
\multirow{6}{*}{Architecture} 
& Input dimension & $p$ (data dimension) \\
& Hidden dimension & 32 \\
& Number of layers & 4 \\
& Time embedding dimension & 128 \\
& Activation & SiLU \\
& Normalisation & LayerNorm \\

\midrule
\multirow{3}{*}{Time Embedding} 
& Embedding type & Learnable embedding \\
& Initialisation & $\mathcal{N}(0, 0.02^2)$ \\
& Conditioning method & Concatenation with $x_t$ \\

\midrule
\multirow{4}{*}{Loss Function} 
& Diffusion loss & MSE \\
& Reconstruction weight ($\lambda_{\mathrm{rec}}$) & 5.0 \\
& Correlation weight ($\lambda_{\mathrm{corr}}$) & 0.1 \\
& Reconstruction loss & MSE \\

\midrule
\multirow{4}{*}{Sampling} 
& Initial distribution & $\mathcal{N}(0, I)$ \\
& Reverse steps & 200 \\
& Noise injection & $\sqrt{\beta_t}\,\epsilon$ \\
& Final projection & Schema-aware (sigmoid + softmax) \\

\midrule
\multirow{4}{*}{Data Representation} 
& Continuous transform & Box-Cox + MinMax $[0,1]$ \\
& Box-Cox $\epsilon$ & $10^{-3}$ \\
& Categorical encoding & One-hot (softmax output) \\
& Binary encoding & 2-dim one-hot \\

\midrule
\multirow{3}{*}{Embeddings} 
& Real embedding size & 1 \\
& Binary embedding size & 2 \\
& Categorical embedding size & 4 \\

\midrule
\multirow{3}{*}{Correlation Loss} 
& Loss type & L1 \\
& Correlation estimate & Batch-wise Pearson \\
& Numerical stability ($\epsilon$) & $10^{-8}$ \\

\bottomrule
\end{tabular}
\caption{Hyperparameters used for training the static tabular diffusion model with schema-aware projection and correlation regularisation.}
\label{tab:diffusion_hyperparameters}
\end{table}

\newpage
\section{Additional Details on Model 4: MCM}\label{App:Model4_All}

We describe a masked clinical modelling (MCM)~\cite{nicholas2025attention} framework for static tabular healthcare data with mixed variable types. The model is designed to reconstruct partially observed patient records and generate synthetic samples via iterative masked refinement. Unlike adversarial or diffusion-based approaches, this method learns conditional dependencies directly through feature masking and reconstruction, while preserving marginal distributions and inter-variable structure.

\subsection{Problem Setup}

Let $X_{\mathrm{real}} \sim P_{\mathrm{data}}$ denote real tabular data, where each observation is a vector:
\[
x \in \mathbb{R}^p.
\]

Let $m \in \{0,1\}^p$ denote a binary mask vector, where:
\[
m_j =
\begin{cases}
1 & \text{if feature } j \text{ is observed}, \\
0 & \text{if feature } j \text{ is masked}.
\end{cases}
\]

The model learns a reconstruction function:
\[
\hat{x} = f_\theta(x, m),
\]
which predicts masked entries conditioned on observed features.

Synthetic samples are generated by iteratively applying masked reconstruction updates:
\[
x^{(t+1)} = m \odot x^{(t)} + (1-m) \odot f_\theta(x^{(t)}, m),
\]
starting from bootstrapped real samples.

\subsection{Symbol Glossary}

\begin{table}[h]
\centering
\begin{tabular}{ll}
\toprule
Symbol & Meaning \\
\midrule
$x$ & Encoded tabular observation \\
$m$ & Feature mask vector \\
$f_\theta$ & Masked reconstruction network \\
$p$ & Total feature dimension (after encoding) \\
$h$ & Hidden representation \\
$\tilde{x}$ & Masked input with mask token applied \\
$\lambda_{\mathrm{corr}}$ & Correlation loss weight \\
$r_{ij}$ & Pearson correlation between variables $i,j$ \\
$R^{\mathrm{real}}$ & Correlation matrix of real data \\
$R^{\mathrm{syn}}$ & Correlation matrix of synthetic data \\
\bottomrule
\end{tabular}
\caption{Notation used in the masked clinical modelling framework.}
\end{table}

\subsection{Data Representation}

The data consist of mixed variable types:
\begin{itemize}
\item Continuous variables, transformed to $[0,1]$ via Box-Cox and min-max scaling,
\item Binary variables represented as two-dimensional one-hot vectors.
\end{itemize}

The encoded feature vector is partitioned as:
\[
x = \big(x_{\mathrm{real}}, x^{(1)}_{\mathrm{bin}}, \dots, x^{(B)}_{\mathrm{bin}}\big),
\]
where $x_{\mathrm{real}} \in [0,1]^{p_r}$ and each binary block $x^{(b)}_{\mathrm{bin}} \in \Delta^{1}$.

A schema object defines the index spans and types of each variable, enabling block-wise processing and type-aware reconstruction.

\subsection{Masked Reconstruction Model}

The model is a neural network:
\[
f_\theta: \mathbb{R}^p \times \{0,1\}^p \rightarrow \mathbb{R}^p,
\]
which predicts full feature vectors from partially masked inputs.

Masked inputs are constructed as:
\[
\tilde{x} = x \odot m + t_{\mathrm{mask}} \odot (1-m),
\]
where $t_{\mathrm{mask}} \in \mathbb{R}^p$ is a learnable mask token.

\subsection{Feature Attention Mechanism}

The model incorporates feature-wise attention layers. Given input $x$ and mask $m$, attention weights are computed as:
\[
a = \mathrm{softmax}(W x),
\]
with masked entries suppressed:
\[
a_j = 0 \quad \text{if } m_j = 0.
\]

The attended representation is:
\[
\tilde{x}_{\mathrm{att}} = a \odot x.
\]

Two such attention stages are used:
\begin{itemize}
\item an input-space attention layer conditioned on the mask,
\item a hidden-space attention layer operating on latent features.
\end{itemize}

\subsection{Network Architecture}

The reconstruction network consists of:
\begin{enumerate}
\item Mask token application to produce $\tilde{x}$,
\item Input-space attention layer,
\item Multi-layer perceptron (MLP) encoder:
\[
h = \mathrm{MLP}_1(\tilde{x}_{\mathrm{att}}),
\]
\item Residual projection from input space:
\[
h = h + \mathrm{ReLU}(W_{\mathrm{skip}} \tilde{x}),
\]
\item Hidden-space attention layer,
\item MLP decoder:
\[
\hat{x}_{\mathrm{raw}} = \mathrm{MLP}_2(h),
\]
\item Schema-aware output projection.
\end{enumerate}

The output is structured as:
\[
x_{\mathrm{real}} = \sigma(\hat{x}_{\mathrm{raw}}^{\mathrm{real}}), \quad
x^{(b)}_{\mathrm{bin}} = \mathrm{softmax}(\hat{x}_{\mathrm{raw}}^{(b)}),
\]
ensuring valid ranges and probability distributions.

\subsection{Reconstruction Objective}

The reconstruction loss is defined only on masked components.

For continuous variables:
\[
\mathcal{L}_{\mathrm{MSE}} =
\|x_{\mathrm{real}} - \hat{x}_{\mathrm{real}}\|_2^2 \quad \text{(masked positions only)}.
\]

For binary variables:
\[
\mathcal{L}_{\mathrm{CE}} =
-\log \hat{p}_{\mathrm{true}},
\]
where $\hat{p}_{\mathrm{true}}$ is the predicted probability of the true class.

The total reconstruction loss is:
\[
\mathcal{L}_{\mathrm{recon}} =
\mathcal{L}_{\mathrm{MSE}} + \mathcal{L}_{\mathrm{CE}}.
\]

\subsection{Correlation Alignment Loss}

To preserve dependency structure, correlation matrices are computed:
\[
R^{\mathrm{real}}, \quad R^{\mathrm{syn}}.
\]

A stochastic clipping is applied:
\[
\tilde{R}_{ij} = \min(u_{ij}, |R_{ij}|)\cdot \mathrm{sign}(R_{ij}), \quad u_{ij} \sim \mathcal{U}(0,1).
\]

The alignment loss is:
\[
\mathcal{L}_{\mathrm{corr}} =
\|\tilde{R}^{\mathrm{syn}} - \tilde{R}^{\mathrm{real}}\|_1.
\]

The final objective is:
\[
\mathcal{L} =
\mathcal{L}_{\mathrm{recon}} + \lambda_{\mathrm{corr}} \, \mathcal{L}_{\mathrm{corr}}.
\]

\subsection{Sampling Procedure}

Synthetic data are generated by iterative masked refinement.

Given a reference dataset $\mathcal{X}_{\mathrm{ref}}$, initial samples are drawn by bootstrap:
\[
x^{(0)} \sim \mathcal{X}_{\mathrm{ref}}.
\]

At each iteration:
\begin{enumerate}
\item Sample a mask $m$ with fixed masking ratio,
\item Compute reconstruction $\hat{x} = f_\theta(x^{(t)}, m)$,
\item Update:
\[
x^{(t+1)} = m \odot x^{(t)} + (1-m)\odot \hat{x}.
\]
\end{enumerate}

After $T$ refinement rounds, the final synthetic sample $x^{(T)}$ is returned.

\subsection{Optimisation}

The model is trained by minimising the reconstruction and correlation losses over masked samples:
\[
\min_\theta \;
\mathbb{E}_{x,m}
\left[
\mathcal{L}_{\mathrm{recon}}(x, m) + \lambda_{\mathrm{corr}} \, \mathcal{L}_{\mathrm{corr}}
\right].
\]

Optimisation is performed using Adam, with masks sampled independently for each minibatch.

\newpage
\subsection{Hyperparameters}\label{App:Model4_Hyperparameters}
\begin{table}[h]
\centering
\small
\begin{tabular}{lll}
\toprule
\textbf{Category} & \textbf{Hyperparameter} & \textbf{Value} \\
\midrule

\multirow{3}{*}{Training} 
& Batch size & 32 \\
& Epochs & 20 \\

\midrule
\multirow{3}{*}{Optimisation} 
& Optimiser & Adam \\
& Learning rate & $1 \times 10^{-3}$ \\
& Adam $(\beta_1, \beta_2)$ & $(0.9, 0.99)$ \\

\midrule
\multirow{4}{*}{Masking (Training)} 
& Min mask ratio & 0.10 \\
& Max mask ratio & 0.95 \\
& Mask sampling & $\mathcal{U}(0.10, 0.95)$ \\
& Mask type & Feature-wise (schema positions) \\

\midrule
\multirow{3}{*}{Masking (Sampling)} 
& Sample mask ratio (default) & 0.50 \\
& Sample mask ratio (used) & 0.25 \\
& Mask type & Fixed ratio \\

\midrule
\multirow{3}{*}{Generation} 
& Number of rounds & 3 \\
& Initialisation & Bootstrap from training data \\
& Sampling method & With replacement \\

\midrule
\multirow{5}{*}{Architecture} 
& Input dimension & $p$ (schema dimension) \\
& Hidden dimension & 128 \\
& Number of layers & 3 (not used) \\
& Dropout & 0.0 \\
& Backbone & Attention $\rightarrow$ MLP $\rightarrow$ Attention $\rightarrow$ MLP \\

\midrule
\multirow{3}{*}{Attention} 
& Type & Linear feature attention \\
& Mask-aware & First layer only \\
& Heads & 1 (implicit) \\

\midrule
\multirow{4}{*}{Loss Function} 
& Reconstruction loss & MSE + Cross-Entropy \\
& Correlation weight ($\lambda_{\mathrm{corr}}$) & 10.0 \\
& Correlation loss & L1 \\
& Masking in loss & Masked positions only \\

\midrule
\multirow{4}{*}{Data Representation} 
& Continuous transform & Box-Cox + MinMax $[0,1]$ \\
& Box-Cox $\epsilon$ & $10^{-3}$ \\
& Binary encoding & 2-dim one-hot \\
& Categorical encoding & None (binary only) \\

\midrule
\multirow{3}{*}{Embeddings} 
& Real embedding size & 1 \\
& Binary embedding size & 2 \\
& Categorical embedding size & 4 (unused) \\

\midrule
\multirow{3}{*}{Correlation Loss} 
& Loss type & L1 \\
& Correlation estimate & Batch-wise Pearson \\
& Regularisation & Random clipping ($\mathcal{U}(0,1)$) \\

\bottomrule
\end{tabular}
\caption{Hyperparameters used for training the MCM framework.}
\label{tab:mcm_hyperparameters}
\end{table}

\newpage
\section{Additional Details on Framework Stream 1: Descriptive Fidelity}\label{App:InfStream1}
\begin{algorithm}[h!]
\caption{Stream 1: Descriptive fidelity evaluation}
\label{alg:descriptive-fidelity}
\begin{algorithmic}[1]
\Require Real dataset $D_{\mathrm{real}}$, trained generator $G$, sample size $n$, transform parameters $\Theta$
\Ensure Marginal, cohort-level, and subgroup-level descriptive comparisons

\State Generate synthetic samples:
\[
D_{\mathrm{syn}} \gets G(n)
\]

\State Decode real and synthetic data to original-scale variables
\[
\widetilde{D}_{\mathrm{real}} \gets \mathrm{Decode}(D_{\mathrm{real}}, \Theta), \qquad
\widetilde{D}_{\mathrm{syn}} \gets \mathrm{Decode}(D_{\mathrm{syn}}, \Theta)
\]

\State \textbf{Marginal distribution similarity}
\For{each variable $X_j$}
    \If{$X_j$ is continuous}
        \State Plot overlaid KDEs for $\widetilde{D}_{\mathrm{real}}^{(j)}$ and $\widetilde{D}_{\mathrm{syn}}^{(j)}$
    \Else
        \State Plot side-by-side normalized histograms for $\widetilde{D}_{\mathrm{real}}^{(j)}$ and $\widetilde{D}_{\mathrm{syn}}^{(j)}$
    \EndIf
\EndFor

\State \textbf{Overall descriptive fidelity}
\State Compute cohort-level Table 1 summaries for real and synthetic data
\State Compare means, standard deviations, medians, interquartile ranges, and category proportions

\State \textbf{Subgroup descriptive fidelity}
\For{each predefined subgrouping variable (\textit{e.g.,} IRSD, age group)}
    \State Partition $\widetilde{D}_{\mathrm{real}}$ and $\widetilde{D}_{\mathrm{syn}}$ into strata
    \For{each stratum}
        \State Compute subgroup-specific Table 1 summaries
        \State Compare real and synthetic descriptive profiles within stratum
    \EndFor
\EndFor

\State \Return descriptive fidelity results across marginal, cohort-level, and subgroup-level evaluations
\end{algorithmic}
\end{algorithm}

\paragraph{Generation and decoding.}
Synthetic samples \(D_{\mathrm{syn}} = G(n)\) are generated and, together with \(D_{\mathrm{real}}\), decoded to original-scale variables \(\widetilde{D}_{\mathrm{syn}}, \widetilde{D}_{\mathrm{real}}\) for interpretable evaluation.

\paragraph{Marginal Distribution similarity.} We compare univariate distributions for each variable. Continuous variables (\textit{e.g.,} age, BMI, HbA1c, eGFR, SBP, time-to-event) are assessed using KDEs, while categorical variables (\textit{e.g.,} IRSD, smoking, comorbidities, outcomes) are compared via normalised histograms. Visualisations are presented in a unified grid.

This provides a baseline check of first-order statistics but does not capture multivariate structure. KDE may introduce boundary artefacts for strictly positive variables due to unbounded kernels.

\paragraph{Overall clinical summary.} We construct a Table~1-style summary of the synthetic cohort. Continuous variables are reported as mean (SD) and median [IQR], while categorical variables are summarised as proportions. The 5-year event rate is included as a cohort-level outcome. This assesses aggregate clinical plausibility but may mask heterogeneity.

\paragraph{Stratified descriptive fidelity.} We evaluate subgroup fidelity via IRSD quintiles and age bands \([30,40), [40,50), [50,60), [60,75)\), computing within-stratum summaries. This tests preservation of subgroup structure beyond pooled marginals, while remaining descriptive in nature.

\newpage
\section{Additional Details on Framework Stream 2: Clinical Utility}\label{App:InfStream2}
\begin{algorithm}[h!]
\caption{Stream 2: Clinical utility evaluation}
\label{alg:clinical-utility}
\begin{algorithmic}[1]
\Require Real dataset $D_{\mathrm{real}}$, synthetic generator $G$, number of synthetic samples $n$, time horizon $\tau$, subgroup definitions $\mathcal{H}$
\Ensure Utility and calibration results for synthetic data

\State Generate synthetic dataset:
\[
D_{\mathrm{syn}} \gets G(n)
\]

\State Prepare real and synthetic datasets for survival modelling:
\[
X_{\mathrm{real}} \gets \mathrm{PrepareCox}(D_{\mathrm{real}}), \qquad
X_{\mathrm{syn}} \gets \mathrm{PrepareCox}(D_{\mathrm{syn}})
\]

\State \textbf{Effect preservation}
\State Fit Cox proportional hazards models on $X_{\mathrm{real}}$ and $X_{\mathrm{syn}}$
\State Extract coefficients, hazard ratios, and confidence intervals
\State Compare covariate effects between real-trained and synthetic-trained models
\State Visualise agreement using side-by-side forest plots

\State \textbf{Global calibration}
\State Compute predicted risk at time horizon $\tau$ for the real dataset under both models
\State Construct calibration curves by binning predicted risks
\State Compute calibration slope and summary calibration error (\textit{i.e.,} D21)

\State \textbf{Subgroup calibration}
\For{each subgroup $h \in \mathcal{H}$}
    \State Restrict the real dataset to subgroup $h$
    \State Compute predicted risk under both models
    \State Construct subgroup-specific calibration curves
    \State Compute subgroup-specific calibration slope and D21
\EndFor

\State Aggregate effect-preservation and calibration results
\State \Return clinical utility assessment of synthetic data
\end{algorithmic}
\end{algorithm}

\paragraph{Generation and preparation.}
Synthetic data \(D_{\mathrm{syn}} = G(n)\) are generated and transformed into Cox-model inputs \(X_{\mathrm{syn}}\), alongside \(X_{\mathrm{real}}\), using identical preprocessing.

\paragraph{Survival-model utility.} We fit Cox proportional hazards models~\cite{cox1972regression} on real and synthetic data using identical covariates. We compare coefficients, hazard ratios (HRs), and confidence intervals to assess preservation of clinically meaningful associations. Discrepancies are quantified via differences in HRs and log-coefficients and visualised using forest plots.

\paragraph{Calibration of predicted risk.} We evaluate calibration by applying both models to the real dataset and computing predicted risk at a fixed horizon. Calibration curves compare predicted and observed event rates across bins. Performance is summarised using the calibration slope and D21 (\(|\text{slope} - 1|\))~\cite{kuo2024ck4gen}.

\paragraph{Stratified calibration.} Calibration is further assessed within IRSD quintiles and age bands. Subgroup-specific curves and D21 quantify whether synthetic-trained models maintain consistent risk estimation across populations.

\newpage
\subsection{Cox Proportional Hazards and Risk Reconstruction}
\label{App:InfStream2_IntroduceCox}

We evaluate downstream clinical utility using Cox proportional hazards (CoxPH) models, which provide a semi-parametric framework for time-to-event modelling.

\subsubsection*{Model formulation}

Let \(X_i \in \mathbb{R}^p\) denote covariates for individual \(i\). The Cox model specifies the hazard:
\[
h_i(t) = h_0(t)\exp(\eta_i), \qquad \eta_i = X_i^\top \beta,
\]
where \(h_0(t)\) is an unspecified baseline hazard and \(\eta_i\) is the linear predictor (LPH). In practice, we use centred predictors \((X_i-\mu)\), yielding
\[
\eta_i = (X_i - \mu)^\top \beta.
\]

\subsubsection*{From hazard to risk}

The survival function is
\[
S_i(t) = [S_0(t)]^{\exp(\eta_i)},
\]
where \(S_0(t)\) is the baseline survival. Predicted risk at horizon \(t\) is
\[
\hat{r}_i(t) = 1 - S_i(t).
\]
Thus, CoxPH produces calibrated probabilities via a monotonic transformation of \(\eta_i\), analogous to a probabilistic output layer.

\subsection{Calibration Metrics and D21}
\label{App:InfStream2_IntroduceMetrics}

We assess whether predicted risks \(\hat{r}_i(t)\) align with observed outcomes using empirical calibration.

\subsubsection*{Empirical calibration curve}

For a fixed time \(t\), define binary outcomes \(y_i(t)\in\{0,1\}\) indicating event occurrence. Individuals are partitioned into \(K\) quantile bins (we use 20 bins) by \(\hat{r}_i(t)\). For each bin \(B_k\),
\[
\bar{\hat{r}}_k = \frac{1}{|B_k|}\sum_{i\in B_k}\hat{r}_i(t), \quad
\bar{y}_k = \frac{1}{|B_k|}\sum_{i\in B_k} y_i(t).
\]
These points approximate the conditional expectation
\[
\mathbb{P}(Y(t)=1 \mid \hat{r}(t)=r) = r.
\]

\subsubsection*{Calibration slope}

We summarise the calibration curve via a regression through the origin:
\[
\bar{y}_k \approx \beta(t)\,\bar{\hat{r}}_k,
\]
yielding slope \(\beta(t)\). Interpretation:
\begin{itemize}
\item \(\beta=1\): perfect calibration,
\item \(\beta<1\): under-dispersion (underprediction),
\item \(\beta>1\): over-dispersion (overprediction).
\end{itemize}

\subsubsection*{D21 miscalibration metric}

We quantify deviation using
\[
D_{21}(t) = |1 - \beta(t)|.
\]
This provides a scalar measure of calibration error, with \(0\) indicating perfect agreement and larger values reflecting increasing miscalibration.

\subsubsection*{Interpretation}

Calibration evaluates absolute risk accuracy rather than ranking. Unlike discrimination metrics, D21 directly measures whether predicted probabilities correspond to observed frequencies, making it critical for clinical decision-making.

\newpage
\section{Additional Details on Framework Stream 3: Structural Validity}\label{App:InfStream3}
\begin{algorithm}[h!]
\caption{Stream 3: Structural validity evaluation}
\label{alg:structural-validity}
\begin{algorithmic}[1]
\Require Real dataset $D_{\mathrm{real}}$, synthetic generator $G$, number of synthetic samples $n$, seed set $\mathcal{S}$
\Ensure Structural validity and stability results for synthetic data

\State Generate synthetic dataset:
\[
D_{\mathrm{syn}} \gets G(n)
\]

\State Prepare real and synthetic datasets for causal discovery:
\[
X_{\mathrm{real}} \gets \mathrm{PrepareGES}(D_{\mathrm{real}}), \qquad
X_{\mathrm{syn}} \gets \mathrm{PrepareGES}(D_{\mathrm{syn}})
\]

\State Remove outcome variables from both datasets
\State Standardize all variables in $X_{\mathrm{real}}$ and $X_{\mathrm{syn}}$

\State \textbf{Graph recovery}
\State Apply GES to $X_{\mathrm{real}}$ and $X_{\mathrm{syn}}$
\State Extract directed edge sets from both recovered graphs
\State Compare graph structures using edge-level and global overlap metrics
\State Visualise recovered graphs for qualitative comparison

\State \textbf{Stability assessment}
\State Define a reference graph from the real dataset
\For{each seed $s \in \mathcal{S}$}
    \State Generate $D_{\mathrm{syn}}^{(s)} \gets G(n; s)$
    \State Prepare and standardize $X_{\mathrm{syn}}^{(s)} \gets \mathrm{PrepareGES}(D_{\mathrm{syn}}^{(s)})$
    \State Apply GES to $X_{\mathrm{syn}}^{(s)}$
    \State Extract directed edges from the recovered graph
    \State Compare recovered structure against the reference graph
    \State Compute SHD and adjacency/orientation precision, recall, and F1
\EndFor

\State Aggregate structural metrics across seeds
\State \Return structural validity assessment of synthetic data
\end{algorithmic}
\end{algorithm}

\paragraph{Generation and preparation.}
Synthetic data \(D_{\mathrm{syn}} = G(n)\) are generated and transformed into \(X_{\mathrm{syn}}\), alongside \(X_{\mathrm{real}}\), using identical preprocessing. Outcome variables are excluded and all features are standardised.

\paragraph{Causal-structure recovery.}
We apply Greedy Equivalence Search (GES) to real and synthetic datasets to recover directed graphs over predictors. Directed edge sets are extracted and compared to assess structural agreement.

We quantify similarity using edge overlap metrics and Jaccard similarity, complemented by visual comparison of recovered graphs. This evaluates whether synthetic data preserve directional relationships beyond marginal distributions.

\paragraph{Stability of structural recovery.}
We assess robustness by repeating synthetic generation across multiple seeds and re-running GES~\cite{chickering2002learning}. A reference graph from real data is used for comparison.

For each run, we compute adjacency and orientation precision, recall, and F1, alongside Structural Hamming Distance (SHD)~\cite{peters2013structural}. Results are summarised across runs to evaluate both fidelity and reproducibility of recovered structure.

\newpage
\subsection{GES: Model, Optimisation, and Implementation Details}
\label{App:InfStream3_IntroduceGES}

We employ Greedy Equivalence Search (GES) for score-based causal discovery under linear--Gaussian assumptions, combining penalised likelihood estimation with greedy search over equivalence classes.

\subsubsection*{Statistical Model}

Let \(X = (X_1,\dots,X_p)^\top\) be generated by an unknown DAG \(\mathcal{G}^*\). For a candidate graph \(\mathcal{G}\),
\[
p(x) = \prod_{i=1}^p p(x_i \mid x_{\mathrm{Pa}_{\mathcal{G}}(i)}),
\quad
X_i = \sum_{j \in \mathrm{Pa}_{\mathcal{G}}(i)} \beta_{ji} X_j + \varepsilon_i,
\]
with \(\varepsilon_i \sim \mathcal{N}(0,\sigma_i^2)\) independent.

Model selection uses the Bayesian Information Criterion:
\[
\mathrm{BIC}(\mathcal{G}) = -2 \log L(\widehat{\theta}_{\mathcal{G}}) + k_{\mathcal{G}} \log n,
\]
providing an implicit sparsity penalty without additional tuning.

\subsubsection*{Equivalence-Class Search}

GES operates over Markov equivalence classes represented as CPDAGs. The optimisation
\[
\widehat{\mathcal{G}} = \arg\min_{\mathcal{G}} \mathrm{BIC}(\mathcal{G})
\]
is approximated via greedy hill-climbing:\\
\hspace*{5mm} (i) forward phase (edge additions), and\\
\hspace*{5mm} (ii) backward phase (edge removals),\\
subject to acyclicity. The output is a locally optimal completed partially directed acyclic graph (CPDAG).

\subsubsection*{Data Representation}

All variables are converted to numeric form: continuous variables are retained, categorical variables are integer-encoded, and outcomes are excluded. Inputs are standardised:
\[
X_j \leftarrow \frac{X_j - \mu_j}{\sigma_j},
\]
ensuring comparable scale and numerical stability.

\subsubsection*{Graph Extraction}

GES returns a CPDAG; we retain only compelled edges when constructing directed graphs. This restricts evaluation to statistically identifiable orientations.

\subsubsection*{Hyperparameters and Implementation}

GES is effectively hyperparameter-free in our setting:
\begin{itemize}
\item BIC score (fixed),
\item sample size \(n\) (full dataset),
\item integer encoding of categorical variables,
\item z-score normalisation,
\item exclusion of outcome variables,
\item fixed random seeds (deterministic search).
\end{itemize}
No constraints on parent set size are imposed.

\subsubsection*{Identifiability}

Under linear--Gaussian assumptions, structure is identifiable only up to Markov equivalence; thus GES recovers an equivalence class rather than a unique DAG.

\newpage
\subsection{Structural Recovery Metrics}
\label{App:InfStream3_StrucMetrics}

We quantify structural fidelity by comparing graphs learned from synthetic data against a reference graph derived from real data, using metrics that capture both adjacency (skeleton) and orientation (directed edge) agreement.

\paragraph{Notation.}
Let \(V = \{1,\dots,p\}\) denote variables, with directed edge sets \(E^\ast\) (reference) and \(\widehat{E}\) (synthetic). The corresponding skeletons are
\[
S^\ast = \operatorname{skel}(E^\ast), \quad \widehat{S} = \operatorname{skel}(\widehat{E}),
\]
where \(\operatorname{skel}(E) = \{\{i,j\} : (i,j)\in E \text{ or } (j,i)\in E\}\).

\paragraph{Adjacency recovery.}
We first evaluate recovery of undirected structure:
\[
\mathrm{TP}_{\mathrm{adj}} = |\widehat{S} \cap S^\ast|,\ 
\mathrm{FP}_{\mathrm{adj}} = |\widehat{S} \setminus S^\ast|,\ 
\mathrm{FN}_{\mathrm{adj}} = |S^\ast \setminus \widehat{S}|.
\]
Precision, recall, and F1 are computed in the standard manner. These metrics assess whether correct variable dependencies are identified, irrespective of direction.

\paragraph{Orientation recovery.}
We next evaluate directed edge agreement:
\[
\mathrm{TP}_{\mathrm{ori}} = |\widehat{E} \cap E^\ast|,\ 
\mathrm{FP}_{\mathrm{ori}} = |\widehat{E} \setminus E^\ast|,\ 
\mathrm{FN}_{\mathrm{ori}} = |E^\ast \setminus \widehat{E}|.
\]
Orientation metrics penalise both missing and incorrectly directed edges, providing a stricter assessment of structural fidelity.

\subsubsection*{Structural Hamming Distance (SHD)}
\label{subsec:shd-math}

\paragraph{Setup.}
Let \(V=\{1,\dots,p\}\) be nodes, with directed edge sets \(E^\ast\) (reference) and \(\widehat{E}\) (estimate). Define skeletons \(S^\ast=\operatorname{skel}(E^\ast)\) and \(\widehat{S}=\operatorname{skel}(\widehat{E})\), where edges are treated as unordered pairs \(\{i,j\}\), \(i<j\).

\paragraph{Definition.}
SHD measures structural discrepancy across unordered node pairs. For each \(\{i,j\}\), we assign a penalty if:
(i) adjacency differs, i.e. \(\{i,j\}\in S^\ast \triangle \widehat{S}\), or
(ii) both graphs contain an edge but with opposite orientation.

Formally,
\[
\mathrm{SHD}(E^\ast,\widehat{E})
=
\sum_{1\le i<j\le p}
d_{ij},
\]
where
\[
d_{ij} =
\begin{cases}
1, & \text{if adjacency differs},\\
1, & \text{if orientations disagree},\\
0, & \text{otherwise}.
\end{cases}
\]

\paragraph{Interpretation.}
SHD penalises missing edges, spurious edges, and reversed directions, providing a unified measure of discrepancy between graph structures.

\newpage
\section{Additional Details on Results: Stream 1}\label{App:MoreResultsStream1}

\subsection{Supplementary Result to Content in Main Text}
\paragraph{Continuous variables.}
Reduced dispersion is accompanied by systematic median shifts. Diffusion consistently overestimates (\textit{e.g.,} Age: \(+10\)–\(+12\) years; BMI: \(+7\)–\(+9\)), indicating structural bias. WGAN-GP underestimates key variables (\textit{e.g.,} SBP up to \(-20.87\) mmHg), while MCM remains closest to the real data.

\paragraph{Categorical variables.}
Disease prevalence differences are small (\(\pm 1\)–\(4\) \%), whereas smoking status shows larger shifts (up to \(\pm 20\) \%), particularly for WGAN-GP and Diffusion.

\begin{table*}[h!]
\centering
\small
\renewcommand{\arraystretch}{1.2}
\caption{IRSD-stratified differences in median values between synthetic and real data.}
\label{tab:irsd-median-delta}

\begin{tabular}{llcccc}
\toprule
\textbf{Characteristic} & \textbf{IRSD} & \textbf{WGAN-GP} & \textbf{Hybrid} & \textbf{Diffusion} & \textbf{MCM} \\
\midrule

\multicolumn{6}{l}{\textbf{Age (years)}} \\
 & 1 & +3.40 & +3.42 & +10.00 & +1.24 \\
 & 2 & +4.54 & +1.29 & +11.38 & +0.85 \\
 & 3 & +5.61 & +4.97 & +12.14 & +2.24 \\
 & 4 & +0.81 & +1.76 & +11.36 & +0.96 \\
 & 5 & +1.14 & +4.98 & +12.06 & +1.07 \\

\midrule
\multicolumn{6}{l}{\textbf{Body Mass Index (kg/m$^2$)}} \\
 & 1 & -1.78 & +1.73 & +7.46 & -2.48 \\
 & 2 & -2.39 & +2.07 & +8.61 & -2.14 \\
 & 3 & -1.70 & +1.46 & +8.59 & -1.49 \\
 & 4 & -1.61 & +0.10 & +8.41 & -1.61 \\
 & 5 & -2.40 & +1.56 & +9.23 & -1.52 \\

\midrule
\multicolumn{6}{l}{\textbf{HbA1c (\%)}} \\
 & 1 & +0.07 & -0.10 & +0.13 & +0.21 \\
 & 2 & +0.17 & -0.18 & +0.21 & +0.18 \\
 & 3 & +0.19 & -0.21 & +0.15 & +0.27 \\
 & 4 & -0.17 & -0.17 & +0.19 & +0.24 \\
 & 5 & +0.00 & -0.01 & +0.18 & +0.34 \\

\midrule
\multicolumn{6}{l}{\textbf{eGFR (mL/min/1.73m$^2$)}} \\
 & 1 & -4.50 & -1.86 & +7.74 & -0.94 \\
 & 2 & -2.34 & +0.48 & +6.62 & -1.69 \\
 & 3 & -4.30 & -1.11 & +7.21 & -1.56 \\
 & 4 & -2.80 & +0.79 & +6.73 & -1.41 \\
 & 5 & -2.40 & -1.94 & +7.27 & -2.08 \\

\midrule
\multicolumn{6}{l}{\textbf{Systolic Blood Pressure (mmHg)}} \\
 & 1 & -15.33 & +1.19 & +10.04 & +1.21 \\
 & 2 & -15.00 & -1.95 & +12.94 & +0.76 \\
 & 3 & -15.00 & +7.37 & +12.57 & +0.82 \\
 & 4 & -11.93 & +0.38 & +12.88 & +0.86 \\
 & 5 & -20.87 & +4.47 & +13.18 & +2.30 \\

\midrule
\multicolumn{6}{l}{\textbf{Time to Event (years)}} \\
 & 1 & +0.08 & -0.06 & +1.05 & -0.02 \\
 & 2 & -0.00 & +0.03 & +1.05 & -0.06 \\
 & 3 & +0.15 & +0.06 & +1.11 & -0.04 \\
 & 4 & +0.11 & +0.19 & +1.08 & -0.07 \\
 & 5 & +0.04 & +0.09 & +1.10 & -0.08 \\

\bottomrule
\end{tabular}
\end{table*}

\newpage

\begin{table}[h!]
\centering
\small
\caption{IRSD-stratified differences in disease prevalence (percentage points).}
\begin{tabular}{llcccc}
\toprule
Characteristic & IRSD & WGAN-GP & Hybrid & Diffusion & MCM \\
\midrule

Diabetes & 1 & +0.02 & +0.01 & +0.00 & -0.01 \\
         & 2 & -0.00 & -0.00 & +0.01 & -0.01 \\
         & 3 & +0.02 & -0.01 & -0.01 & -0.04 \\
         & 4 & -0.01 & -0.00 & +0.04 & -0.00 \\
         & 5 & -0.01 & +0.02 & +0.03 & -0.01 \\

CKD & 1 & -0.01 & -0.00 & +0.01 & -0.00 \\
    & 2 & -0.01 & +0.00 & +0.01 & -0.00 \\
    & 3 & -0.01 & +0.01 & +0.00 & +0.00 \\
    & 4 & -0.01 & -0.00 & +0.02 & +0.00 \\
    & 5 & -0.00 & +0.00 & +0.04 & -0.00 \\

AF & 1 & -0.01 & -0.01 & -0.01 & -0.00 \\
   & 2 & -0.01 & -0.01 & -0.01 & -0.00 \\
   & 3 & -0.01 & -0.01 & -0.01 & -0.00 \\
   & 4 & -0.01 & -0.01 & -0.00 & -0.00 \\
   & 5 & -0.01 & -0.01 & -0.00 & -0.00 \\

CVD Event & 1 & -0.00 & +0.01 & +0.01 & -0.01 \\
          & 2 & -0.03 & -0.00 & +0.03 & -0.01 \\
          & 3 & -0.02 & +0.02 & -0.00 & -0.01 \\
          & 4 & -0.02 & +0.01 & +0.03 & -0.01 \\
          & 5 & -0.01 & +0.02 & +0.03 & -0.01 \\

\bottomrule
\end{tabular}
\end{table}

\begin{table}[h!]
\centering
\small
\caption{IRSD-stratified differences in smoking status proportions (percentage points).}
\begin{tabular}{llcccc}
\toprule
Characteristic & IRSD & WGAN-GP & Hybrid & Diffusion & MCM \\
\midrule

Non-smoker & 1 & +0.19 & +0.10 & -0.20 & +0.03 \\
           & 2 & +0.21 & -0.07 & -0.11 & +0.09 \\
           & 3 & +0.16 & +0.09 & +0.05 & +0.09 \\
           & 4 & +0.15 & +0.01 & -0.27 & +0.03 \\
           & 5 & +0.14 & +0.12 & -0.22 & +0.04 \\

Ex-smoker & 1 & -0.20 & -0.06 & +0.26 & -0.02 \\
          & 2 & -0.18 & +0.10 & +0.15 & -0.02 \\
          & 3 & -0.16 & -0.07 & -0.02 & -0.04 \\
          & 4 & -0.15 & +0.03 & +0.27 & +0.01 \\
          & 5 & -0.14 & -0.09 & +0.22 & -0.02 \\

Current smoker & 1 & +0.01 & -0.04 & -0.06 & -0.01 \\
               & 2 & -0.03 & -0.03 & -0.04 & -0.07 \\
               & 3 & +0.00 & -0.02 & -0.03 & -0.05 \\
               & 4 & +0.00 & -0.04 & -0.00 & -0.04 \\
               & 5 & +0.00 & -0.02 & +0.01 & -0.02 \\

\bottomrule
\end{tabular}
\end{table}

\newpage
\subsection{Additional Results: Age-Stratified Distributional Fidelity}
\label{sec:appendix_age_strat}

To complement the IRSD-stratified analysis, we evaluate marginal fidelity under an alternative conditioning axis by stratifying on age groups. These results assess whether the subgroup-level discrepancies previously observed persist across different partitions of the data.

\paragraph{Continuous variables.}
Table~\ref{tab:age-median-delta} mirrors the IRSD-stratified findings, showing that reduced dispersion is again accompanied by systematic median shifts. Diffusion exhibits consistent overestimation across age groups (\textit{e.g.,} BMI \(+8\)–\(+10\)), while WGAN-GP underestimates key variables such as systolic blood pressure (up to \(-21.60\) mmHg). MCM remains closest to the real data with comparatively small deviations. These patterns align with Section~\ref{Sec:Results}, confirming that subgroup-level biases are not specific to IRSD stratification.

\begin{table*}[h!]
\centering
\small
\renewcommand{\arraystretch}{1.2}
\caption{Age-group–stratified differences in median values between synthetic and real data.}
\label{tab:age-median-delta}

\begin{tabular}{llcccc}
\toprule
\textbf{Characteristic} & \textbf{Age Group} & \textbf{WGAN-GP} & \textbf{Hybrid} & \textbf{Diffusion} & \textbf{MCM} \\
\midrule

\multicolumn{6}{l}{\textbf{Age (years)}} \\
 & [30,40) & +0.02 & -0.70 & -0.26 & +0.02 \\
 & [40,50) & +0.20 & +0.57 & +1.52 & +0.68 \\
 & [50,60) & +0.53 & -0.04 & +2.80 & -2.98 \\
 & [60,75) & -0.36 & -1.45 & -1.51 & -0.29 \\

\midrule
\multicolumn{6}{l}{\textbf{Body Mass Index (kg/m$^2$)}} \\
 & [30,40) & -1.59 & +0.60 & +10.53 & -1.68 \\
 & [40,50) & -1.84 & +1.08 & +9.33 & -1.91 \\
 & [50,60) & -1.99 & +2.16 & +8.53 & -1.62 \\
 & [60,75) & -2.62 & +0.42 & +8.39 & -1.74 \\

\midrule
\multicolumn{6}{l}{\textbf{HbA1c (\%)}} \\
 & [30,40) & -0.31 & -0.42 & -0.20 & +0.26 \\
 & [40,50) & -0.06 & -0.24 & +0.06 & +0.26 \\
 & [50,60) & +0.18 & +0.05 & +0.13 & +0.23 \\
 & [60,75) & +0.25 & -0.25 & +0.11 & +0.17 \\

\midrule
\multicolumn{6}{l}{\textbf{eGFR (mL/min/1.73m$^2$)}} \\
 & [30,40) & +0.43 & +2.12 & +9.22 & -3.59 \\
 & [40,50) & -1.64 & +0.65 & +8.81 & -2.32 \\
 & [50,60) & -3.40 & -3.04 & +8.71 & -0.53 \\
 & [60,75) & -3.50 & +1.83 & +9.72 & +1.17 \\

\midrule
\multicolumn{6}{l}{\textbf{Systolic Blood Pressure (mmHg)}} \\
 & [30,40) & -21.60 & -3.29 & +3.26 & +6.19 \\
 & [40,50) & -17.95 & -0.12 & +6.48 & +2.82 \\
 & [50,60) & -14.98 & +3.04 & +8.76 & -0.51 \\
 & [60,75) & -14.82 & +1.94 & +7.00 & -4.55 \\

\midrule
\multicolumn{6}{l}{\textbf{Time to Event (years)}} \\
 & [30,40) & +0.24 & +0.26 & +0.91 & -0.07 \\
 & [40,50) & +0.11 & +0.16 & +1.00 & -0.07 \\
 & [50,60) & +0.01 & -0.07 & +1.07 & -0.06 \\
 & [60,75) & +0.01 & +0.06 & +1.14 & -0.02 \\

\bottomrule
\end{tabular}
\end{table*}

\newpage
\paragraph{Categorical variables.}
Tables~\ref{tab:age-disease-delta} and~\ref{tab:age-smoking-delta} show that disease prevalence deviations remain small across age groups (typically within a few percentage points), whereas smoking status exhibits larger discrepancies, particularly for WGAN-GP and Diffusion. This is consistent with the IRSD-based results, indicating that behavioural variables are more difficult to preserve than clinical endpoints.

\begin{table}[h!]
\centering
\small
\renewcommand{\arraystretch}{1.2}
\caption{Differences in disease prevalence and cardiovascular event rates across age groups (percentage points).}
\label{tab:age-disease-delta}
\begin{tabular}{llcccc}
\toprule
\textbf{Characteristic} & \textbf{Age Group} & \textbf{WGAN-GP} & \textbf{Hybrid} & \textbf{Diffusion} & \textbf{MCM} \\
\midrule

\multicolumn{6}{l}{\textbf{Diabetes}} \\
& [30,40) & -0.02 & -0.02 & +0.03 & -0.01 \\
& [40,50) & -0.00 & -0.03 & +0.03 & -0.01 \\
& [50,60) & +0.02 & +0.04 & -0.02 & -0.02 \\
& [60,75) & -0.03 & -0.03 & -0.05 & -0.03 \\

\midrule
\multicolumn{6}{l}{\textbf{CKD}} \\
& [30,40) & -0.00 & -0.00 & +0.01 & +0.00 \\
& [40,50) & -0.00 & -0.00 & +0.02 & +0.00 \\
& [50,60) & -0.01 & +0.01 & +0.00 & -0.00 \\
& [60,75) & -0.01 & -0.01 & +0.01 & -0.00 \\

\midrule
\multicolumn{6}{l}{\textbf{AF}} \\
& [30,40) & -0.00 & -0.00 & +0.01 & -0.00 \\
& [40,50) & -0.00 & -0.00 & -0.00 & +0.00 \\
& [50,60) & -0.01 & -0.01 & -0.01 & -0.00 \\
& [60,75) & -0.01 & -0.01 & -0.01 & -0.01 \\

\midrule
\multicolumn{6}{l}{\textbf{CVD Event}} \\
& [30,40) & -0.00 & -0.01 & +0.02 & -0.00 \\
& [40,50) & -0.00 & -0.01 & +0.03 & +0.00 \\
& [50,60) & -0.01 & +0.03 & +0.00 & -0.01 \\
& [60,75) & -0.07 & +0.02 & -0.04 & -0.03 \\

\bottomrule
\end{tabular}
\end{table}

\begin{table}[h!]
\centering
\small
\renewcommand{\arraystretch}{1.2}
\caption{Differences in smoking status proportions across age groups (percentage points).}
\label{tab:age-smoking-delta}
\begin{tabular}{llcccc}
\toprule
\textbf{Characteristic} & \textbf{Age Group} & \textbf{WGAN-GP} & \textbf{Hybrid} & \textbf{Diffusion} & \textbf{MCM} \\
\midrule

\multicolumn{6}{l}{\textbf{Non-smoker}} \\
& [30,40) & +0.17 & +0.03 & -0.06 & +0.06 \\
& [40,50) & +0.16 & -0.06 & -0.14 & +0.02 \\
& [50,60) & +0.17 & +0.06 & -0.10 & +0.06 \\
& [60,75) & +0.20 & +0.21 & -0.14 & +0.08 \\

\midrule
\multicolumn{6}{l}{\textbf{Ex-smoker}} \\
& [30,40) & -0.17 & -0.01 & +0.10 & -0.03 \\
& [40,50) & -0.17 & +0.06 & +0.15 & +0.01 \\
& [50,60) & -0.17 & -0.03 & +0.12 & -0.03 \\
& [60,75) & -0.17 & -0.11 & +0.16 & -0.04 \\

\midrule
\multicolumn{6}{l}{\textbf{Current smoker}} \\
& [30,40) & -0.00 & -0.02 & -0.04 & -0.03 \\
& [40,50) & +0.01 & +0.00 & -0.01 & -0.04 \\
& [50,60) & -0.00 & -0.03 & -0.02 & -0.03 \\
& [60,75) & -0.03 & -0.10 & -0.03 & -0.04 \\

\bottomrule
\end{tabular}
\end{table}

\newpage
\paragraph{Joint distribution consistency.}
Beyond marginal distributions, Table~\ref{tab:age-irsd-composition-delta} evaluates the composition of IRSD quintiles within age groups. All models exhibit shifts in subgroup representation, with Diffusion showing pronounced overrepresentation of disadvantaged strata in younger age groups. These deviations indicate that the joint distribution \(P(\text{Age}, \text{IRSD})\) is not preserved, extending the subgroup inconsistencies observed in Section~\ref{Sec:Results}.

\begin{table*}[h!]
\centering
\small
\renewcommand{\arraystretch}{1.2}
\caption{Differences in IRSD quintile composition within age groups (percentage points). Positive values indicate overrepresentation in the synthetic data relative to the real data; negative values indicate underrepresentation.}
\label{tab:age-irsd-composition-delta}
\begin{tabular}{llcccc}
\toprule
\textbf{Characteristic} & \textbf{Age Group} & \textbf{WGAN-GP} & \textbf{Hybrid} & \textbf{Diffusion} & \textbf{MCM} \\
\midrule

\multicolumn{6}{l}{\textbf{IRSD Q1}} \\
& [30,40) & -0.12 & +0.04 & +0.53 & +0.03 \\
& [40,50) & -0.12 & -0.03 & +0.33 & +0.00 \\
& [50,60) & -0.10 & +0.00 & +0.08 & +0.02 \\
& [60,75) & -0.12 & +0.03 & -0.02 & +0.02 \\

\midrule
\multicolumn{6}{l}{\textbf{IRSD Q2}} \\
& [30,40) & +0.11 & +0.00 & -0.13 & -0.03 \\
& [40,50) & +0.12 & +0.07 & -0.10 & -0.03 \\
& [50,60) & +0.15 & +0.03 & -0.07 & -0.01 \\
& [60,75) & +0.19 & -0.14 & -0.08 & -0.02 \\

\midrule
\multicolumn{6}{l}{\textbf{IRSD Q3}} \\
& [30,40) & -0.07 & -0.05 & -0.22 & -0.01 \\
& [40,50) & -0.08 & -0.10 & -0.16 & -0.02 \\
& [50,60) & -0.05 & -0.06 & +0.06 & +0.01 \\
& [60,75) & +0.01 & +0.05 & +0.15 & -0.00 \\

\midrule
\multicolumn{6}{l}{\textbf{IRSD Q4}} \\
& [30,40) & +0.11 & +0.07 & -0.07 & -0.04 \\
& [40,50) & +0.09 & +0.16 & -0.02 & -0.06 \\
& [50,60) & +0.04 & +0.06 & -0.02 & -0.04 \\
& [60,75) & +0.01 & +0.01 & -0.03 & -0.05 \\

\midrule
\multicolumn{6}{l}{\textbf{IRSD Q5}} \\
& [30,40) & -0.02 & -0.06 & -0.11 & +0.05 \\
& [40,50) & -0.02 & -0.10 & -0.05 & +0.10 \\
& [50,60) & -0.04 & -0.03 & -0.05 & +0.02 \\
& [60,75) & -0.09 & +0.04 & -0.02 & +0.05 \\

\bottomrule
\end{tabular}
\end{table*}

\newpage
\section{Additional Details on Results: Stream 2}\label{App:MoreResultsStream2}

To complement the clinical utility analysis in Section~\ref{Sec:Results}, we provide detailed hazard ratio (HR) estimates across all covariates. These results assess whether the apparent agreement observed in Figure~\ref{fig:4ClinicalUtility}(a) holds consistently across variables and model classes.

\paragraph{Continuous covariates.}
While GAN-based methods and MCM broadly align with real-data HRs, this agreement is variable-specific. WGAN-GP preserves some effects (\textit{e.g.,} Age) but introduces distortions in others (\textit{e.g.,} eGFR direction reversal), while the Hybrid model exhibits systematic inflation (\textit{e.g.,} Age, BMI). MCM remains the most consistent, preserving both direction and magnitude across covariates. In contrast, Diffusion produces substantial deviations, including exaggerated effects (\textit{e.g.,} BMI) and sign reversals (\textit{e.g.,} HbA1c, eGFR), indicating instability beyond simple scaling differences.

\paragraph{Categorical covariates.}
Categorical effects follow a similar pattern. Smoking-related covariates exhibit large discrepancies, with WGAN-GP attenuating effects (\textit{e.g.,} ex-smoker) and Hybrid inflating them. Diffusion again produces extreme and implausible HRs (\textit{e.g.,} ex-smoker \(>20\)), while MCM remains comparatively stable. IRSD effects are also distorted, particularly for Diffusion, which yields highly inflated and inconsistent estimates across quintiles.

\paragraph{Consistency with calibration.}
These results provide a mechanistic explanation for the calibration patterns observed in Section~\ref{Sec:Results}. Models with more stable HR estimates (\textit{e.g.,} WGAN-GP, MCM) achieve better population-level calibration, whereas large deviations and directional errors (as seen in Diffusion) lead to miscalibrated risk predictions, particularly across subgroups.

\begin{table}[h!]
\centering
\footnotesize
\caption{Hazard Ratios for Continuous Covariates in Real and Synthetic PRIME-CVD Data.}
\begin{tabular}{lccccc}
\toprule
\textbf{Covariate} & \textbf{Real} & \textbf{WGAN-GP} & \textbf{Hybrid} & \textbf{MCM} & \textbf{Diffusion} \\
\midrule

Age 
& 1.07 [1.07, 1.08]
& 1.07 [1.06, 1.08]
& 1.12 [1.11, 1.13]
& 1.06 [1.06, 1.07]
& 1.32 [1.31, 1.33] \\

HbA1c 
& 2.00 [1.95, 2.06]
& 1.90 [1.83, 1.98]
& 1.75 [1.71, 1.78]
& 2.08 [2.00, 2.17]
& 0.50 [0.46, 0.55] \\

BMI 
& 1.03 [1.02, 1.04]
& 1.02 [1.00, 1.03]
& 1.04 [1.03, 1.05]
& 1.05 [1.03, 1.06]
& 1.68 [1.65, 1.71] \\

eGFR 
& 0.99 [0.98, 0.99]
& 1.02 [1.01, 1.03]
& 0.98 [0.98, 0.98]
& 0.97 [0.96, 0.98]
& 0.56 [0.56, 0.57] \\

SBP 
& 1.01 [1.01, 1.02]
& 1.02 [1.02, 1.02]
& 1.01 [1.01, 1.01]
& 1.03 [1.02, 1.03]
& 1.05 [1.04, 1.05] \\

\bottomrule
\end{tabular}
\end{table}

\begin{table}[h!]
\centering
\scriptsize
\caption{Hazard Ratios for Smoking Status in Real and Synthetic PRIME-CVD Data.}
\begin{tabular}{lccccc}
\toprule
\textbf{Covariate} & \textbf{Real} & \textbf{WGAN-GP} & \textbf{Hybrid} & \textbf{MCM} & \textbf{Diffusion} \\
\midrule

Smoking (current) 
& 1.52 [1.33, 1.74]
& 1.64 [1.07, 2.52]
& 1.41 [1.17, 1.69]
& 1.52 [1.22, 1.88]
& 0.66 [0.53, 0.81] \\

Smoking (ex) 
& 1.44 [1.29, 1.61]
& 0.80 [0.67, 0.96]
& 2.32 [2.13, 2.52]
& 1.31 [1.13, 1.51]
& 23.68 [21.14, 26.52] \\

\bottomrule
\end{tabular}
\end{table}

\begin{table}[h!]
\centering
\scriptsize
\caption{Hazard Ratios for IRSD Quintiles in Real and Synthetic PRIME-CVD Data (Reference: Quintile 5).}
\begin{tabular}{lccccc}
\toprule
\textbf{Covariate} & \textbf{Real} & \textbf{WGAN-GP} & \textbf{Hybrid} & \textbf{MCM} & \textbf{Diffusion} \\
\midrule

IRSD quintile 1 
& 1.01 [0.88, 1.15]
& 0.95 [0.78, 1.16]
& 1.02 [0.92, 1.13]
& 0.98 [0.83, 1.15]
& 3.89 [2.70, 5.61] \\

IRSD quintile 2 
& 1.12 [0.97, 1.29]
& 0.97 [0.76, 1.22]
& 0.80 [0.70, 0.90]
& 1.26 [1.04, 1.53]
& 0.31 [0.20, 0.48] \\

IRSD quintile 3 
& 1.17 [1.03, 1.33]
& 0.63 [0.52, 0.77]
& 0.99 [0.89, 1.12]
& 1.29 [1.10, 1.51]
& 8.60 [6.03, 12.27] \\

IRSD quintile 4 
& 1.03 [0.89, 1.19]
& 0.54 [0.43, 0.67]
& 0.89 [0.77, 1.03]
& 0.75 [0.61, 0.91]
& 219.91 [153.48, 315.11] \\

\bottomrule
\end{tabular}
\end{table}

\end{document}